\documentclass[10pt,journal,compsoc]{IEEEtran}

\ifCLASSOPTIONcompsoc
  \usepackage[nocompress]{cite}
\else
  \usepackage{cite}
\fi

\ifCLASSINFOpdf
  \usepackage[pdftex]{graphicx}
  \usepackage{multirow}
  \usepackage{bbding}
  \usepackage{amsfonts}
\else
\fi

\usepackage{amsmath}
\usepackage{pifont}

\usepackage{amsmath,amsfonts}
\usepackage{array}
\usepackage[caption=false,font=normalsize,labelfont=sf,textfont=sf]{subfig}
\usepackage{textcomp}
\usepackage{stfloats}
\usepackage{url}
\usepackage{verbatim}
\usepackage{graphicx}
\usepackage{amsmath,amssymb}
\usepackage[ruled,vlined,linesnumbered]{algorithm2e}
\SetKwInOut{KwInit}{Initialize}
\usepackage{cite}
\usepackage{pifont}
\usepackage{tikz}
\usepackage{setspace}
\usepackage{booktabs}
\usepackage{multirow}
\usepackage{makecell}
\usepackage{multirow}
\usepackage{graphicx}
\usepackage{booktabs}
\usepackage{xcolor}
\usepackage{graphicx}
\usepackage{booktabs}
\usepackage[table]{xcolor}
\usepackage{graphicx}
\usepackage{multirow}
\usepackage{colortbl}
\usepackage{amssymb}
\usepackage{graphicx}
\usepackage[
  colorlinks=true,
  linkcolor=red, 
  citecolor=blue,
  urlcolor=blue,
]{hyperref}

\usepackage[
justification=justified,singlelinecheck=false
]{caption} 

\definecolor{lightgrayrow}{RGB}{235,240,247}

\definecolor{bestred}{RGB}{220,20,60}
\definecolor{secondblue}{RGB}{30,90,200}
\definecolor{oursgray}{RGB}{245,245,245}

\newcommand{\best}[1]{\textcolor{bestred}{\textbf{#1}}}
\newcommand{\second}[1]{\textcolor{secondblue}{#1}}

\newcommand{\circled}[1]{%
  \tikz[baseline=(char.base)]{
    \node[draw,circle,inner sep=0.3pt] (char) {\scriptsize #1};
  }%
}
\usepackage{enumitem}
\usepackage{ragged2e}
\newcommand{\ourMethod}{\textit{O2D-MedNet}\@\xspace} 
\definecolor{ltgray}{rgb}{0.84,0.84,0.84}
\definecolor{ltblue}{rgb}{0.356,0.608,0.835}
\usepackage[normalem]{ulem}

\begin{document}
\title{O2D-MedNet: An Optimization-Aware Dual-Domain Framework for Robust Medical Image Segmentation}

\author{Shipeng Hu, 
Miaohui Wang,~
Jing Qin,~
Sukun Tian,
and Yuchun Sun

\IEEEcompsocitemizethanks{ 
\IEEEcompsocthanksitem Shipeng Hu, Yuchun Sun and Sukun Tian are with the Center of Digital Dentistry, Peking University School and Hospital of Stomatology. Miaohui Wang is with the State Key Laboratory of Radio Frequency Heterogeneous Integration and Guangdong Key Laboratory of Intelligent Information Processing, Shenzhen University. 
Jing Qin is with the School of Nursing, Hong Kong Polytechnic University.
}
}

\markboth{O2D-MedNet: An Optimization-Aware Dual-Domain Framework for Robust Medical Image Segmentation}%
{Shell \MakeLowercase{\textit{et al.}}: Bare Advanced Demo of IEEEtran.cls for IEEE Computer Society Journals}

\IEEEtitleabstractindextext{%
\begin{abstract}\justifying
Medical image segmentation (MIS) remains challenging due to composite degradations and multi-scale geometric variations, which jointly impair lesion-background separability, structural consistency, and optimization stability. Existing single-stream frameworks entangle heterogeneous cues within a single representation pathway, while hybrid-stream approaches often rely on generic or static fusion, limiting effective coordination across domains and failing to explicitly address optimization instability. 
To overcome these limitations, we propose a new optimization-aware dual-domain framework for MIS (\ourMethod). First, a degradation-aware  dual-domain interaction (DDI) module integrates \textit{Fourier}- and \textit{Wavelet}-domain features  to suppress degradation-sensitive responses while preserving structural and fine-detail information. Second, a cross-backbone contour-guided fusion (CGF) module enables selective interaction between CNN-based local detail modeling and visual state-space-based global structure modeling, improving boundary accuracy and representation consistency. Third, a gradient-context optimization (GCO) strategy stabilizes training by adaptively adjusting learning dynamics based on loss trends and gradient statistics, mitigating instability caused by class imbalance and heterogeneous sample difficulty. Extensive experiments on six benchmarks demonstrate consistent improvements over state-of-the-art MIS methods across multiple imaging modalities. Further analyses confirm enhanced sensitivity, structural consistency, and optimization stability, validating the effectiveness of the proposed \ourMethod{} for robust MIS under heterogeneous conditions.
\end{abstract}

}

\maketitle

\IEEEdisplaynontitleabstractindextext
\IEEEpeerreviewmaketitle

\section{Introduction}
\label{sec:intro}
\IEEEPARstart{M}{edical} image segmentation (MIS)~\cite{qiu2023rethinking,zhang2024structure} is a fundamental task in medical image analysis, aiming to assign semantic labels to pixels or voxels for accurate delineation of lesions or anatomical structures. Robust MIS supports a wide range of clinical applications \cite{antonelli2022medical,liu2026brainparc}, including \textit{disease screening}, \textit{diagnosis}, \textit{treatment planning}, and \textit{longitudinal assessment}. 
However, the increasing volume and diversity of medical imaging data (\textit{i.e.}, \textit{optical}, \textit{radiological}, and \textit{ultrasound} images) continue to impose substantial demands  on clinical workflows, where manual inspection remains time-consuming and error-prone. Consequently, developing robust and scalable MIS frameworks has become increasingly important for enabling reliable clinical deployment.

Despite recent advances, achieving robust MIS performance across multiple organs and imaging conditions remains challenging \cite{yang2023self}. Specifically, medical images often exhibit composite degradations (\textit{e.g.}, \textit{noise}, \textit{artifacts}, and \textit{specular reflections}) together with multi-scale geometric variability in lesion and anatomical structure \textit{size}, \textit{shape}, and \textit{location}. These factors can jointly degrade lesion–background separability, disrupt structural consistency, and lead to unstable optimization during model training. As a result, MIS approaches often struggle to simultaneously preserve fine-grained  boundaries, maintain global coherence, and generalize across heterogeneous imaging samples.
\begin{figure}[!t]
\centering
\includegraphics[width=0.98\linewidth]{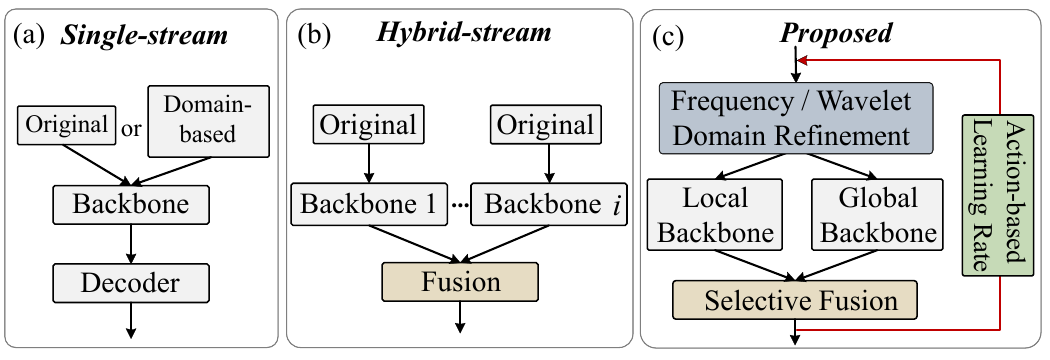}
\caption{Comparison of representative medical image segmentation (MIS) paradigms. (a) Single-stream frameworks process either original or domain-based inputs through a single  pathway, limiting their ability to disentangle heterogeneous cues~\cite{medley2021cycoseg,huang2022missformer,nam2024modality}. (b) Hybrid-stream frameworks introduce multiple backbone branches and fuse their features to exploit complementary representations, but the interaction is typically generic or static~\cite{zhang2021transfuse,he2023h2former}. (c) In contrast, \ourMethod{} integrates dual-domain refinement, explicit local–global decoupling, contour-guided selective fusion, and optimization-aware learning in a unified framework.}
\label{fig1}
\end{figure}

Existing MIS methods can be broadly categorized into single-stream and hybrid-stream frameworks, as illustrated in \textcolor{blue}{Figs.}~\ref{fig1} (a) and (b). Single-stream approaches rely on a dominant representation pathway for feature extraction and prediction. Representative CNN-based models~\cite{medley2021cycoseg,zhang2025eag} are effective at capturing local textures and boundary details, whereas Transformer- and state-space-based architectures~\cite{huang2022missformer,wu2022fat} improve long-range dependency modeling and global structure aggregation. Some studies further incorporate frequency-domain priors~\cite{zhou2024gobletnet,nam2024modality} to enhance robustness against image degradations. 
Hybrid-stream frameworks extend this idea by introducing multiple representation pathways, such as CNN-Transformer combinations~\cite{zhang2021transfuse,he2023h2former}, frequency-decomposition architectures~\cite{zhou2023xnet}, and task-specific auxiliary streams~\cite{qiu2023rethinking,han2026mambaeviscrib}. Although these methods improve  representation diversity, they generally rely on static feature fusion strategies or heuristic interactions, limiting their ability to adapt to local details, global structures, and degradation-sensitive responses under complex conditions.
\begin{table*}[!t]
\centering
\caption{Taxonomy and comparison of representative medical image segmentation (MIS) frameworks.
Existing methods are categorized according to their dominant representation and stream architecture, with emphasis on local-global modeling, domain knowledge, cross-stream interaction, degradation modeling, and optimization awareness.}
\resizebox{\linewidth}{!}{
\begin{tabular}{lcccccccc}
\hline
\hline
\multirow{2}{*}{{Method}} &\multirow{2}{*}{{Category}} &\multirow{2}{*}{{Representation}} &{Local} &{Global} &{Domain} &{Cross-stream} &{Degradation} &{Optimization} \\ 
&&&Detail   &Context  &Knowledge  &Interaction  &Modeling   &Awareness \\ \hline

Reference Baselines\\
\textit{DeepLabv3+}~\cite{chen2018encoder} & Single & Atrous CNN + ASPP & Explicit & Explicit & No & No & No & No \\
\textit{U$^2$-Net}~\cite{qin2020u2} & Single & Nested U-shaped CNN & Explicit & Explicit & No & No & No & No \\
\textit{MedSAM}~\cite{ma2024segment} &Hybrid & ViT + Box Prompt & Implicit & Explicit & No & Explicit & No & No \\

\hline
Single-stream Frameworks\\

\textit{ACNN}~\cite{oktay2017anatomically} &Single &CNN + Anatomical Prior &Explicit &Limited &Anatomical Prior &No &Limited &No \\

\textit{DRINet}~\cite{chen2018drinet} &Single &U-shaped CNN &Explicit &Limited &No &No &No &No \\

\textit{BCDU-Net}~\cite{azad2019bi} &Single &U-shaped + BiConvLSTM &Explicit &Limited &No &No &No &No \\

\textit{CyCoSeg}~\cite{medley2021cycoseg} &Single &CNN + Shape Prior &Explicit &Limited &Shape Prior &Limited &No &No \\

\textit{Swin-UNet}~\cite{cao2022swin} &Single &Transformer &Explicit &Explicit &No &No &No &No \\

\textit{MISSFormer}~\cite{huang2022missformer} &Single &Transformer &Explicit &Explicit &No &No &No &No \\

\textit{Mamba-UNet}~\cite{wang2024mamba} &Single &VSSM &Implicit &Explicit &No &No &No &No \\

\textit{LKM-UNet}~\cite{wang2024lkm} &Single &VSSM &Explicit &Explicit &No &No &No &No \\

\textit{Mamba-Sea}~\cite{cheng2025mamba} &Single &VSSM &Explicit &Explicit &No &No &Limited &No \\

\textit{GobletNet}~\cite{zhou2024gobletnet} &Single &Wavelet-enhanced CNN &Explicit &Limited &Wavelet &Explicit &Limited &No \\

\textit{MADGNet}~\cite{nam2024modality} &Single &Multi-frequency Attention &Explicit &Limited &Multi-frequency (DCT) &No &Limited &No \\

\textit{FFTMed}~\cite{pham2025fftmed} &Single &Fourier U-shaped &Limited &Explicit &Fourier &No &Explicit &No \\

\textit{SAM2-UNet}\cite{xiong2026sam2} & Single & SAM2-Hiera + U-Net & Explicit & Explicit & No & No & No & No \\

\hline
Hybrid-stream Frameworks\\

\textit{MedT}~\cite{valanarasu2021medical} &Hybrid &Local-Global Transformer &Explicit &Explicit &No &Limited &No &No \\

\textit{TransFuse}~\cite{zhang2021transfuse} &Hybrid &CNN + Transformer &Explicit &Explicit &No &Explicit &No &No \\

\textit{H2Former}~\cite{he2023h2former} &Hybrid &CNN + Transformer &Explicit &Explicit &No &Explicit &No &No \\

\textit{XNet}~\cite{zhou2023xnet} &Hybrid &Low + High Frequency &Explicit &Limited &Wavelet &Explicit &Limited &No \\

\textit{DS2F}~\cite{qiu2023rethinking} &Hybrid &Segmentation + Super-resolution &Explicit &Limited &No &Explicit &No &No \\

\textit{MambaEviScrib}~\cite{han2026mambaeviscrib} &Hybrid &CNN + Mamba &Explicit &Explicit&No &Explicit &Limited &Limited \\

\hline
\rowcolor{black!16}
\textit{Ours} &Hybrid &CNN + VSSM &Explicit &Explicit &Fourier + Wavelet &Explicit &Explicit &Explicit \\

\hline
\hline
\end{tabular}
}
\label{table:related_works}
\end{table*}

We argue that a fundamental limitation of existing MIS frameworks lies in the coupled treatment of degradation suppression, representation learning, and optimization. In most existing MIS  approaches, these factors are implicitly handled within a unified learning process, making it difficult to dynamically balance feature restoration, local-global representation coordination, and optimization stability. Consequently, MIS performance often deteriorates in the presence of composite degradations, large anatomical variations, class imbalance, and heterogeneous sample difficulty.

To address these challenges, we introduce a new \underline{o}ptimization-aware \underline{d}ual-\underline{d}omain framework for MIS (\ourMethod{}). The proposed method follows a progressive design principle that explicitly decouples degradation suppression, representation learning, and optimization dynamics. First, a dual-domain interaction module exploits complementary \textit{Fourier}- and \textit{Wavelet}-domain information to enhance feature discriminability. Second, a contour-guided cross-backbone fusion mechanism facilitates selective integration of local and global representations, mitigating redundant and noisy feature propagation. Third, a gradient-context-aware optimization strategy dynamically adapts learning behavior according to loss trends and gradient statistics, improving optimization stability under heterogeneous data conditions. Extensive experiments on six benchmark segmentation datasets spanning three medical imaging categories as well as diverse lesions and anatomical structures demonstrate the effectiveness and  generalization of the proposed MIS framework.

The main contributions are summarized as follows:
\begin{itemize}
\item We propose \ourMethod{}, an optimization-aware dual-domain cross-backbone interaction framework that explicitly decouples degradation suppression, representation learning, and training dynamics, enabling robust MIS under composite degradations and multi-scale  anatomical variations.

\item We introduce a degradation-aware dual-domain interaction (DDI) mechanism that exploits complementary \textit{Fourier}- and \textit{Wavelet}-domain representations to improve lesion–background separability while preserving both structural integrity and fine-grained details.

\item We design a cross-backbone contour-guided fusion (CGF) module that adaptively integrates CNN-based local features with state-space-based global representations, enhancing structural consistency across complex anatomical morphologies and scale variations.

\item We develop a stable gradient-context optimization (GCO) strategy that modulates optimization behavior according to loss dynamics and gradient statistics, improving optimization stability and training robustness under class imbalance and heterogeneous sample difficulty.
\end{itemize}

\section{{Related Work}}
\label{sec:related_work}
MIS has been widely studied, with recent advances largely driven by deep neural networks.  From the perspective of network architecture, existing MIS methods can be broadly categorized into \textit{reference baselines}, \textit{single-stream frameworks}, and \textit{hybrid-stream frameworks}, as summarized in \textcolor{blue}{Table}~\ref{table:related_works}.

\subsection{Reference Baselines}
We first review several representative reference baselines that provide distinct architectural paradigms for image segmentation. \textit{DeepLabv3+}~\cite{chen2018encoder} integrates atrous spatial pyramid pooling with an encoder--decoder architecture to capture multi-scale contextual information and progressively refine object boundaries. \textit{U$^2$-Net}~\cite{qin2020u2} adopts a nested U-shaped architecture composed of residual U-blocks, enabling the joint modeling of fine-grained local details and multi-scale contextual features. In contrast, \textit{MedSAM}~\cite{ma2024segment} exemplifies the promptable medical foundation-model paradigm. It employs a ViT-based image encoder to extract image representations and a prompt encoder to embed input prompts, which are subsequently integrated by the mask decoder to generate segmentation masks. These methods represent three complementary architectural paradigms, ranging from conventional encoder-decoder networks and nested multi-scale architectures to prompt-conditioned foundation MIS models.

\subsection{Single-stream Frameworks}
Single-stream MIS frameworks \cite{shi2015semi,rahman2023ambiguous,fischer2026progressive} rely on a dominant network pathway, where feature extraction, context aggregation, and prediction are conducted on a single backbone architecture, including \textit{U-shaped}- or \textit{CNN}-based, \textit{Transformer}-based, \textit{visual state-space model} (VSSM)-based, and other  \textit{domain}-based models.

\noindent \textbf{\textit{U-shaped} or \textit{CNN}-based.}
Early MIS methods are primarily based on U-shaped encoder–decoder or CNN-based networks, where hierarchical convolutional features are progressively encoded and decoded for dense prediction \cite{azad2024medical,kreitner2024synthetic}. Representative models, such as \textit{ACNN}  \cite{oktay2017anatomically}, \textit{DRINet} \cite{chen2018drinet}, \textit{BCDU-Net} \cite{azad2019bi}, \textit{CyCoSeg} \cite{medley2021cycoseg},  \textit{FusionNet} \cite{quan2021fusionnet}, UNeXt \cite{valanarasu2022unext}, \textit{MBP-SSNet} \cite{wang2024multi}, and \textit{EAG-SAE-ESTF} \cite{zhang2025eag}, enhance this paradigm through improved skip connections, contextual modeling, and feature refinement. These methods are effective at capturing local textures and boundary details, which are essential for segmenting small lesions and thin anatomical structures. However, their convolution-dominant design inherently relies on local receptive fields, making it difficult to model long-range dependencies and maintain global structural coherence when anatomical variations are large or spatially dispersed.

\noindent \textbf{\textit{Transformer}-based.}
To address the above limitation, Transformer-based single-stream MIS models introduce self-attention mechanisms to enhance global structure modeling. For instance, \textit{Swin-UNet} \cite{cao2022swin}, \textit{MISSFormer} \cite{huang2022missformer}, \textit{LeViT-UNet} \cite{xu2023levit}, \textit{FAT-Net} \cite{wu2022fat}, and \textit{CSWin-UNet} \cite{liu2025cswin} capture long-range dependencies through various attention designs. More recently, SAM2-UNet~\cite{xiong2026sam2} integrates the Hiera backbone of SAM2 with a U-shaped decoder and lightweight adapters for prompt-free segmentation, retaining a dominant single-stream pathway.  While these approaches improve the representation of large-scale structures, they still operate within a single pathway, where local detail preservation and global structure modeling are tightly coupled. This coupling makes it difficult to adaptively prioritize boundary-sensitive cues or global semantics under varying imaging conditions. Especially, global aggregation may smooth fine structures or weaken subtle lesion boundaries in low-contrast or noisy regions.

\noindent \textbf{\textit{VSSM}-based.}
More recently, VSSM-based architectures, such as \textit{Mamba-UNet} \cite{wang2024mamba}, \textit{LKM-UNet} \cite{wang2024lkm}, \textit{H-vmuNet} \cite{wu2025h}, \textit{UltraLight VM} \cite{wu2025ultralight}, and \textit{Mamba-Sea} \cite{cheng2025mamba}, provide an alternative mechanism for efficient long-range modeling. These methods improve computational efficiency while maintaining global  structure awareness. However, similar to CNN- and Transformer-based MIS schemes, they still encode local details, global structures, and scale variations within a single pathway. Consequently, they continue to face an inherent trade-off between local sensitivity and global coherence, especially under composite degradations and heterogeneous anatomical variations.

\noindent \textbf{\textit{Domain}-based.}
Another line of MIS work incorporates frequency-domain  information to enhance robustness. For example, \textit{GobletNet} \cite{zhou2024gobletnet} explores wavelet-based high-frequency, while \textit{MADGNet} \cite{nam2024modality} and \textit{FFTMed} \cite{pham2025fftmed} introduces multi-frequency of discrete cosine transform (DCT)- or \textit{Fourier}-domain cues to strengthen boundary sensitivity and suppress noise. Although these approaches improve degradation robustness, they typically inject domain information into a single-stream backbone with fixed or weakly adaptive fusion strategies. Consequently, degradation suppression is not explicitly coordinated with representation learning or training dynamics, limiting their effectiveness under complex medical imaging  conditions.

\noindent \textbf{\textit{Summary}.}
Despite their diverse network architectures, single-stream MIS methods fundamentally entangle degradation suppression, representation learning, and optimization within a unified or single pathway. This implicit coupling assumes that a single representation can simultaneously resolve degradation-induced ambiguity, capture multi-scale structural variations, or maintain stable optimization dynamics. However, such an assumption becomes weak or invalid under heterogeneous imaging conditions, where these factors impose competing and condition-dependent requirements. As a result, single-stream MIS models lack the flexibility to adaptively prioritize local detail, global structure, or robustness to degradations, leading to compromised segmentation performance.
\begin{figure*}[!t]
\centering
\includegraphics[width=0.98\linewidth, height=0.352\linewidth]{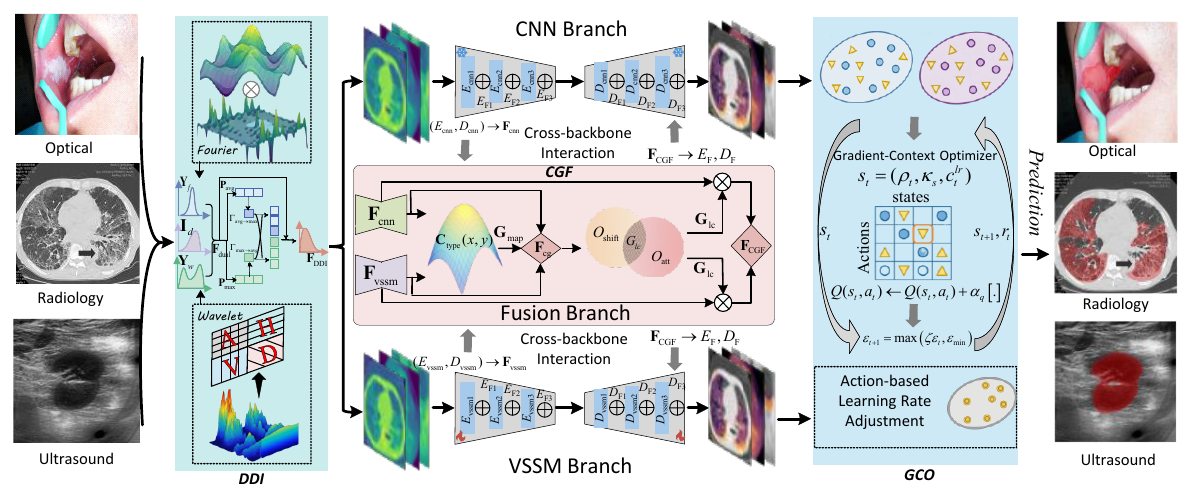}
\caption{Overview of the proposed dual-domain cross-backbone interaction framework for MIS (\ourMethod{}). A dual-domain refinement (DDI) module integrates global spectral and local multi-resolution cues to suppress degradations and enhance feature separability. Parallel CNN and VSSM branches then capture local details and global structures, respectively. A contour-guided fusion (CGF) module enables selective cross-backbone interaction in skip connections to improve multi-scale structural consistency and reduce redundant propagation. Finally, a gradient-context optimization (GCO) strategy adapts training based on loss evolution and gradient statistics to improve optimization stability and generalization.}
\label{fig2}
\end{figure*}

\subsection{Hybrid-stream Frameworks}
Hybrid-stream MIS frameworks \cite{zimmer2019towards,seidlitz2022robust} aim to alleviate the bottlenecks of single-stream models by introducing multiple pathways to exploit complementary information, including \textit{hybrid-backbone}-based, \textit{hybrid-frequency}-based, and \textit{hybrid-task}-based models.

\noindent \textbf{\textit{Hybrid-backbone}-based.}
One common MIS architecture design combines CNN-based local modeling with Transformer-style global modeling. For example, \textit{MedT} \cite{valanarasu2021medical}, \textit{TransFuse}~\cite{zhang2021transfuse}, and \textit{H2Former}~\cite{he2023h2former}  capture both fine-grained local details and long-range dependencies, demonstrating that local and global representations are inherently complementary for MIS. However, these methods implicitly assume that complementary representations can be effectively combined through static or heuristic fusion mechanisms. In addition, they also lack an explicit strategy to determine when local details should dominate and when global structures should be emphasized under varying degradation levels and  boundary ambiguities. Consequently, fusion may propagate redundant or noisy features, or suppress subtle yet important structures.

\noindent \textbf{\textit{Hybrid-frequency}-based.}
Hybrid frequencies \cite{valanarasu2021kiu,zhou2024prototype} provide another form of complementary modeling. For instance, \textit{XNet}~\cite{zhou2023xnet} separates low- and high-frequency components, processes them with dedicated encoders, and produces dual-branch segmentation predictions. While this design improves representation diversity, its complementarity is largely constrained to frequency decomposition and consistency learning. Moreover, it does not explicitly coordinate local-global backbone biases, nor does it address optimization instability during training.

\noindent \textbf{\textit{Hybrid-task}-based.}
Other hybrid MIS approaches incorporate auxiliary tasks or supervision strategies \cite{oksuz2020deep,zhou2022semantic,xie2024integrating}, such as super-resolution \textit{DS2F}~\cite{qiu2023rethinking} or weak supervision~\textit{MambaEviScrib} \cite{han2026mambaeviscrib}. These methods further highlight the benefit of multi-stream modeling under specific settings. Nevertheless, their interaction mechanisms are often task-dependent, and their optimization processes remain governed by fixed or empirically tuned schedules. Consequently, their robustness may degrade under heterogeneous imaging samples or complex degradations.

\noindent \textbf{\textit{Summary}.}
Hybrid-stream MIS frameworks partially relax the representation bottleneck by introducing multiple pathways, yet they typically treat degradation handling, cross-stream interaction, and optimization as loosely coupled components. Their fusion strategies often rely on static or weakly adaptive mechanisms, implicitly assuming that complementary representations can be combined without considering condition-dependent dominance or degradation sensitivity. Moreover, learning optimization remains external to network architectures, limiting the ability to stabilize training. As a result, although hybrid-stream models improve representation diversity, they still lack a unified and adaptive mechanism to coordinate degradation suppression, representation interaction, and optimization dynamics under challenging conditions.

\section{Methodology}
\label{sec:methodology}
The overall architecture of \ourMethod{} is illustrated in \textcolor{blue}{Fig.}~\ref{fig2}. The framework consists of three main components: (1) a degradation-aware dual-domain interaction (DDI) module, (2) a cross-backbone contour-guided fusion (CGF) module, and (3) a stable gradient-context optimization (GCO) strategy. Given an input medical image, DDI first enhances feature discriminability through complementary \textit{Fourier}- and \textit{Wavelet}-domain processing. The refined features are subsequently encoded by parallel CNN and VSSM branches, where CGF enables selective fusion between local and global representations. During training, GCO dynamically adapts optimization behavior according to gradient context to improve optimization  stability.

\subsection{Degradation-Aware Dual-Domain Interaction}
In MIS, composite degradations are often amplified in early feature representations, leading to ambiguous lesion–background boundaries and unstable downstream learning \cite{lu2025wavelet}. Since such degradation-induced responses cannot be reliably suppressed by representation learning alone, we first introduce a dual-domain refinement stage to improve feature separability before high-level modeling.

\subsubsection{Dual-Domain Feature Representation}
Specifically, we propose the DDI module, which jointly exploits  \textit{Fourier}-domain and \textit{Wavelet}-domain representations.  Given the input feature map after a 1$\times$1 convolution, $\mathbf{I}_d \in \mathbb{R}^{B \times C \times H \times W}$, where $B$, $C$, $H$, and $W$ denote the batch size, channel number, height, and width, respectively. The \textit{Fourier} branch performs global  spectral transform and a learnable  filter $(\mathbf{W}_r,\mathbf{W}_i)$ is used to modulate the associated spectrum, yielding: 
\begin{equation}
\small
\mathbf{Y}_f
=
\Re\!\left(
\mathcal{F}^{-1}\!\left(
\mathcal{F}(\mathbf{I}_d)\odot(\mathbf{W}_r+i\mathbf{W}_i)
\right)
\right),
\label{eq:freq_branch}
\end{equation} 
where $\mathbf{Y}_f$ denotes the \textit{Fourier}-domain response after spectral modulation, $\Re(\cdot)$ extracts the real part, and $\odot$ denotes element-wise multiplication. $\mathcal{F}$ and $\mathcal{F}^{-1}$ denote the forward and inverse \textit{Fourier} transforms, respectively. $\mathbf{W}_r$ and $\mathbf{W}_i$ represent the learnable real- and imaginary- part filters used to multiply with $\mathcal{F}(\mathbf{I}_d)$.  The \textit{Fourier} branch mainly suppresses unstable high-frequency responses while maintaining low-frequency structural consistency.

In parallel, the \textit{Wavelet} branch extracts localized multi-resolution features through wavelet decomposition and reconstruction:
\begin{equation}
\small
\mathbf{Y}_w
=
\mathcal{W}^{-1}\!\left(
\mathcal{S}\!\left(\mathcal{W}(\mathbf{I}_d)\right)
\right),
\label{eq:wavelet_branch}
\end{equation}
where $\mathbf{Y}_w$ denotes the \textit{Wavelet}-domain response reconstructed from the modulated multi-scale coefficients. $\mathcal{W}$ and $\mathcal{W}^{-1}$ denote the wavelet decomposition and reconstruction, respectively. $\mathcal{S}(\cdot)$ denotes a learnable subband-wise coefficient selection and modulation operation, which suppresses degradation-sensitive weak coefficients and enhances boundary-related subbands across scales.  The \textit{Wavelet}  branch mainly focuses on multi-resolution local-detail preservation.

\begin{figure*}[!t]
\centering
\includegraphics[width=0.98\linewidth, height=0.30\linewidth]{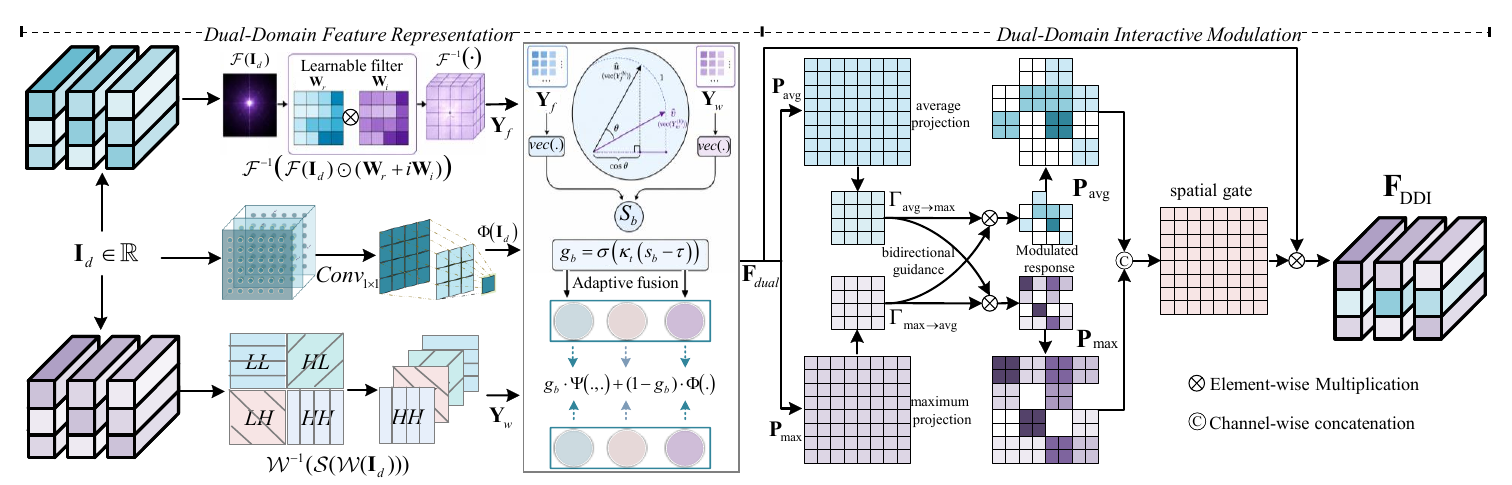}
\caption{Illustration of the proposed DDI module. Input features are processed in parallel by \textit{Fourier} and \textit{Wavelet} branches to capture global structural and multi-resolution detail cues, respectively. Similarity-aware gating adaptively fuses the responses, followed by cross-branch interaction to enhance spatial complementarity.} 
\label{fig3}
\end{figure*}

To adaptively balance \textit{Fourier}-domain global spectral feature and \textit{Wavelet}-domain localized multi-resolution feature, we vectorize $\mathbf{Y}_f$ and $\mathbf{Y}_w$, and compute a sample-wise cosine similarity:
\begin{equation}
\small
s_b
=
\frac{
\left\langle
\texttt{vec}\!\left(\mathbf{Y}_f^{(b)}\right),
\texttt{vec}\!\left(\mathbf{Y}_w^{(b)}\right)
\right\rangle
}{
\left\|
\texttt{vec}\!\left(\mathbf{Y}_f^{(b)}\right)
\right\|_2
\left\|
\texttt{vec}\!\left(\mathbf{Y}_w^{(b)}\right)
\right\|_2+\epsilon_0
},
\quad
b=1,\ldots,B .
\label{eq:cosine_similarity}
\end{equation}
where $s_b$ measures the response consistency between the \textit{Fourier} and \textit{Wavelet} branches for the $b$-th sample, $b$ indexes the sample within the mini-batch with the size of $B$,  $\epsilon_0$ represents a small constant for numerical stability, and $\texttt{vec}(\cdot)$ flattens the feature map into a vector. A larger $s_b$ indicates  stronger agreement between the two domain responses.

Moreover, we construct a gating mechanism $g_b$ to control the sample-wise contribution of the dual-domain response and the original feature mapping: 
\begin{equation}
\small
g_b = \sigma_s\!\left(\kappa_t(s_b-\tau)\right), 
\end{equation}
where $\sigma_s(\cdot)$ denotes  the \texttt{Sigmoid} function, $\kappa_t$ represents a temperature coefficient, and $\tau$=0.5 represents a threshold. As a result, the dual-domain features are fused by 
\begin{equation}
\small
{\mathbf{F}_{\mathrm{dual}}^{(b)}} = g_b\cdot \Psi\!\left(\mathbf{Y}_f^{(b)},\mathbf{Y}_w^{(b)}\right) + (1-g_b)\cdot \Phi\!\left(\mathbf{I}_d^{(b)}\right),
\label{eq:gated_fusion}
\end{equation}
where ${\mathbf{F}_{\mathrm{dual}}^{(b)}}$ denotes the adaptively fused dual-domain feature of the $b$-th sample, and the mini-batch feature ${\mathbf{F}_{\mathrm{dual}}}$ is obtained by stacking all sample-wise outputs $
\mathbf{F}_{\mathrm{dual}}=[\mathbf{F}_{\mathrm{dual}}^{(1)},\mathbf{F}_{\mathrm{dual}}^{(2)},...,\mathbf{F}_{\mathrm{dual}}^{(B)}]$ along the batch dimension. 
 $\Psi(\cdot,\cdot)$ denotes a lightweight fusion mapping for the two domain responses, implemented by channel concatenation followed by a 1$\times$1 convolution. $\Phi(\cdot)$ is implemented by a 1$\times$1 convolution and maps $\mathbf{I}_d^{(b)}$ to the same feature shape as the fused dual-domain response.
As a result, $\Psi(\mathbf{Y}_f^{(b)},\mathbf{Y}_w^{(b)})$ and $\Phi(\mathbf{I}_d^{(b)})$ have compatible dimensions for weighted fusion.

\subsubsection{Dual-Domain Interactive Modulation}
The sample-level gate and the subsequent spatial recalibration address different levels of degradation uncertainty, as shown in \textcolor{blue}{Fig.}~\ref{fig3}. Meanwhile, composite degradations often vary substantially across samples: some images are dominated by specular highlights, whereas others are mainly affected by noise, artifacts, or weak contrast. Therefore, the scalar gate $g_b$ estimates the global reliability of dual-domain responses for each sample, which is broadcast to all channel and spatial positions during fusion.  
However, lesions and anatomical structures are also spatially non-uniform within the same sample. A spatial modulation step is thus required to suppress local false responses while preserving informative lesion regions. In this sense, $g_b$ determines how much dual-domain information should be trusted globally, whereas the spatial gate refines where such information should be emphasized locally.

In the DDI module, we further introduce dual-domain dynamic interaction to enhance spatial complementarity. Specifically,  channel-wise average projection and maximum projection followed by spatial downsampling are applied to ${\mathbf{F}_{\mathrm{dual}}}$, producing two complementary descriptor maps: ${\mathbf{P}_{\mathrm{avg}}}=\texttt{Downsample}(\texttt{Pool}_{\text{avg}}({\mathbf{F}_{\mathrm{dual}}}))$ and ${\mathbf{P}_{\mathrm{max}}}=\texttt{Downsample}(\texttt{Pool}_{\text{max}}({\mathbf{F}_{\mathrm{dual}}}))$, 
where ${\mathbf{P}_{\mathrm{avg}}}$ denotes the average-response descriptor and captures stable region-level responses, while ${\mathbf{P}_{\mathrm{max}}}$ denotes the max-response descriptor and highlights locally discriminative activations and boundary-sensitive cues. $\{{\mathbf{P}_{\mathrm{avg}}}, {\mathbf{P}_{\mathrm{max}}}\} \in\mathbb{R}^{B\times1\times h\times w}$, where $h$ and $w$ denote the downsampled spatial resolution.

We then adopt two zero-initialized convolutional branches to predict dual-domain interactive modulation terms ${\Gamma_{\mathrm{avg}\rightarrow\mathrm{max}}}$ and ${\Gamma_{\mathrm{max}\rightarrow\mathrm{avg}}}$, and perform bidirectional guidance as:
\begin{equation}
\resizebox{0.90\hsize}{!}{$\left({\widehat{\mathbf{P}}_{\mathrm{max}}},{\widehat{\mathbf{P}}_{\mathrm{avg}}}
\right)
=
\left(
\sigma_s\!\bigl({\Gamma_{\mathrm{avg}\rightarrow\mathrm{max}}}({\mathbf{P}_{\mathrm{avg}}})\bigr)\odot{\mathbf{P}_{\mathrm{max}}},
\sigma_s\!\bigl({\Gamma_{\mathrm{max}\rightarrow\mathrm{avg}}}({\mathbf{P}_{\mathrm{max}}})\bigr)\odot {\mathbf{P}_{\mathrm{avg}}}
\right)$},
\label{eq:bidirectional_guidance}
\end{equation}
where ${\widehat{\mathbf{P}}_{\mathrm{max}}}$ and ${\widehat{\mathbf{P}}_{\mathrm{avg}}}$ denote the modulated max-response and average-response descriptors. ${\Gamma_{\mathrm{avg}\rightarrow\mathrm{max}}}(\cdot)$ and ${\Gamma_{\mathrm{max}\rightarrow\mathrm{avg}}}(\cdot)$ denote learnable modulation functions implemented by zero-initialized convolutional branches. Both branches preserve the spatial size of their inputs, so that ${\Gamma_{\mathrm{avg}\rightarrow\mathrm{max}}}({\mathbf{P}_{\mathrm{avg}}})$ and ${\Gamma_{\mathrm{max}\rightarrow\mathrm{avg}}}({\mathbf{P}_{\mathrm{max}}})$ are compatible with ${\mathbf{P}_{\mathrm{max}}}$ and ${\mathbf{P}_{\mathrm{avg}}}$, respectively.   During training, the bidirectional modulation is progressively learned, allowing regional responses to suppress spurious local peaks while preserving boundary-sensitive cues.

Finally, ${\widehat{\mathbf{P}}_{\mathrm{max}}}$ and ${\widehat{\mathbf{P}}_{\mathrm{avg}}}$ are concatenated and transformed into a spatial gate through a 1$\times$1 convolution, bilinear upsampling, and \texttt{Sigmoid} normalization:
\begin{equation}
\small
\resizebox{0.90\hsize}{!}{$\mathbf{F}_{\mathrm{DDI}}
=
\texttt{ReLU}\!\left({\mathbf{F}_{\mathrm{dual}}}
\odot
\sigma_s\!\left(
\texttt{Upsample}\!\left(
\texttt{Conv}_{1\times1}
\left([{\widehat{\mathbf{P}}_{\mathrm{avg}}},{\widehat{\mathbf{P}}_{\mathrm{max}}}]\right)
\right)
\right)
\right)$}.
\label{eq:dd_dio_output}
\end{equation}
where $\mathbf{F}_{\mathrm{DDI}}$ denotes the final output feature of the DDI module. Through this dual-domain interactive modulation, degradation-induced response is effectively reduced before entering the representation stage. This not only improves lesion–background separability, but also provides a more stable foundation for subsequent local–global modeling.

\subsection{Cross-Backbone Contour-Guided Fusion}
After dual-domain feature interaction, the model proceeds to representation learning through parallel CNN and VSSM branches. It is noted that CNN features are sensitive to local textures and boundary details, and VSSM features are effective for modeling long-range dependencies and global semantics. However, directly fusing these heterogeneous representations assumes feature compatibility, which is often violated under complex degradations and scale variations \cite{huang2025acm}.

\begin{figure*}[!t]
\centering
\includegraphics[width=0.92\linewidth, height=0.352\linewidth]{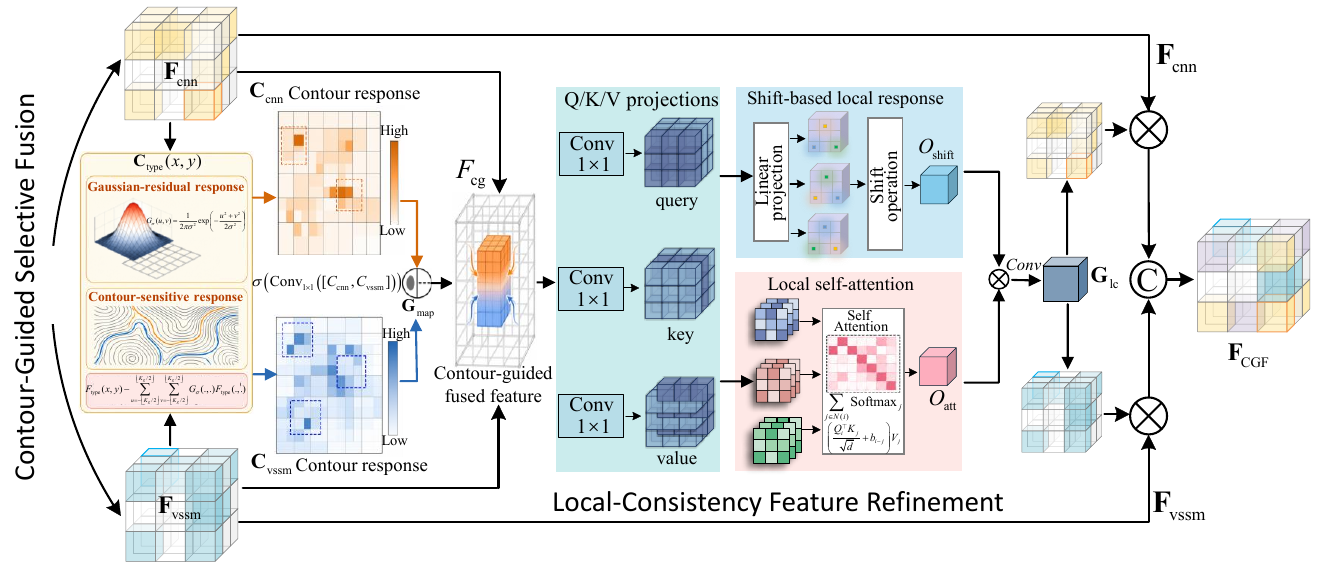}
\caption{Illustration of the CGF module. CNN and VSSM features are transformed into contour-enhanced responses using Gaussian-based high-pass filtering. A contour-guided gating map enables selective cross-backbone fusion, balancing local details and global structures. The fused features are reinjected into both branches and refined by lightweight local attention to enhance multi-scale structural consistency and suppress redundant responses.}
\label{fig4}
\end{figure*}

\subsubsection{Contour-Guided Selective Fusion}
To address this issue, we propose the CGF module as shown in {\color{blue}Fig.~\ref{fig4}},
which performs selective and structure-aware cross-backbone interaction.
The DDI-refined feature $\mathbf{F}_{\mathrm{DDI}}$ is first delivered to the CNN and VSSM branch, where each branch contains three encoder stages and three decoder stages.
The hierarchical feature representations extracted from the CNN and VSSM
encoder-decoder pathways at each corresponding network stage are denoted as
$\mathbf{F}_{\mathrm{cnn}}$ and $\mathbf{F}_{\mathrm{vssm}}$, respectively.
Specifically, the CNN pathway produces stage-wise features through its
encoder stages $(E_{\mathrm{cnn}}^{1},E_{\mathrm{cnn}}^{2},E_{\mathrm{cnn}}^{3})$
and decoder stages $(D_{\mathrm{cnn}}^{1},D_{\mathrm{cnn}}^{2},D_{\mathrm{cnn}}^{3})$,
while the VSSM pathway generates corresponding stage-wise features through
$(E_{\mathrm{vssm}}^{1},E_{\mathrm{vssm}}^{2},E_{\mathrm{vssm}}^{3})$ and
$(D_{\mathrm{vssm}}^{1},D_{\mathrm{vssm}}^{2},D_{\mathrm{vssm}}^{3})$.
At each stage, the paired CNN and VSSM features are fed into CGF for
cross-backbone interaction. Given the paired features $\mathbf{F}_{\mathrm{cnn}}$ and $\mathbf{F}_{\mathrm{vssm}}$, CGF first computes Gaussian-residual responses to emphasize
contour-sensitive transitions in the feature space.
For a spatial location $(x,y)$, this operation is defined as:
\begin{equation}
\small
\begin{gathered}
\mathbf{C}_{\mathrm{type}}(x,y)
=
\mathbf{F}_{\mathrm{type}}(x,y)
-
\sum\nolimits_{u=-\lfloor K_g/2\rfloor}^{\lfloor K_g/2\rfloor}\cdot \sum\nolimits_{v=-\lfloor K_g/2\rfloor}^{\lfloor K_g/2\rfloor},\\
G_{\sigma}(u,v)\mathbf{F}_{\mathrm{type}}(x-u,y-v),
\mathrm{type}\in\{\text{cnn},\text{vssm}\},
\end{gathered}
\label{eq:contour_response}
\end{equation}
where $\mathbf{C}_{\mathrm{type}}$ denotes the contour-sensitive  Gaussian-residual response of the CNN or VSSM branch, $G_{\sigma}$ denotes a normalized Gaussian kernel with standard deviation $\sigma$=1.0, and $K_g$=5 denotes the kernel size. The Gaussian-residual response explicitly suppresses smooth background components while highlighting boundary-relevant structures, thus producing more stable contour-aware responses.

These two contour responses $\mathbf{C}_{\mathrm{cnn}}$ and $\mathbf{C}_{\mathrm{vssm}}$ are then concatenated along the channel dimension, and a spatial gating map ${\mathbf{G}_{\mathrm{map}}}$ is generated:
\begin{equation}
\small
{\mathbf{G}_{\mathrm{map}}}
=
\sigma_s\!\left(
\texttt{Conv}_{1\times1}\!\left([\mathbf{C}_{\text{cnn}},\mathbf{C}_{\text{vssm}}]\right)
\right)
\in
(0,1)^{B\times1\times H\times W}.
\label{eq:spatial_gate}
\end{equation}
where ${\mathbf{G}_{\mathrm{map}}}$ adaptively indicates whether each spatial position should rely more on CNN-based local details or VSSM-based global structure. $[\cdot,\cdot]$ denotes channel-wise concatenation. Consequently, the two backbone features are selectively fused by the  gating map as: 
\begin{equation}
\small
{\mathbf{F}_{\mathrm{cg}}}
=
{\mathbf{G}_{\mathrm{map}}}\odot\mathbf{F}_{\text{cnn}}
+
(1-{\mathbf{G}_{\mathrm{map}}})\odot\mathbf{F}_{\text{vssm}},
\label{eq:selective_fusion}
\end{equation}
where ${\mathbf{F}_{\mathrm{cg}}}$ denotes the contour-guided fused feature. It serves as a carrier for estimating local-consistency cues, rather than as the final output of CGF.

\subsubsection{Local-Consistency Feature Refinement}
On top of the contour-guided fusion, ${\mathbf{F}_{\mathrm{cg}}}$ is projected by three 1$\times$1 convolutions to generate the query $\mathbf{Q}$, key $\mathbf{K}$, and value $\mathbf{V}$. The upper path performs shift-based local feature refinement. Specifically, the projected feature is first processed by a linear projection and then shifted along neighboring spatial directions to exchange local responses. The shifted features are concatenated and projected by a 1$\times$1 convolution to obtain the shift-based local response ${\mathbf{O}_{\mathrm{shift}}}$:
\begin{equation}
\small
\resizebox{0.88\hsize}{!}{$\mathbf{O}_{\mathrm{shift}}
=
\texttt{Conv}_{1\times1}
\left(
\texttt{Concat}_{r\in\mathcal{R}}
\left(
\texttt{Shift}_{r}
\left(
\phi_{\mathrm{shift}}\big([\mathbf{Q},\mathbf{K},\mathbf{V}]\big)
\right)
\right)
\right)$},
\label{eq:shift_interaction}
\end{equation}
where $[\mathbf{Q},\mathbf{K},\mathbf{V}]$ denotes channel-wise concatenation of the query, key, and value features. 
$\phi_{\mathrm{shift}}(\cdot)$ denotes a linear projection for channel mixing, 
$\mathcal{R}$ denotes the set of spatial shift directions, and 
$\operatorname{Shift}_{r}(\cdot)$ shifts the projected feature along direction $r$.

In parallel, the lower path computes local self-attention within a local window $\mathcal{N}(i)$:
\begin{equation}
\small
\mathbf{O}_{\mathrm{att}}(i)
=
\sum_{j\in\mathcal{N}(i)}
\frac{
\texttt{exp}\!\left(\frac{\mathbf{Q}_i^\top \mathbf{K}_j}{\sqrt{d}}+b_{i-j}\right)
}{
\sum_{k\in\mathcal{N}(i)}
\texttt{exp}\!\left(\frac{\mathbf{Q}_i^\top \mathbf{K}_k}{\sqrt{d}}+b_{i-k}\right)
}
\mathbf{V}_j ,
\label{eq:local_attention}
\end{equation}
where $\mathbf{O}_{\mathrm{att}}$ denotes the local self-attention response, $i$ denotes the query position after flattening the spatial dimension of the feature map, and $j$ and $k$ denote neighboring positions in $\mathcal{N}(i)$, $d$ denotes the feature dimension, and $b_{i-j}$ denotes the relative positional bias.  The shift-based path captures displacement-aware local refinement, while the self-attention path aggregates context-aware neighborhood responses.

The shift-based response and the local self-attention response are then combined to estimate a local-consistency modulation map ${\mathbf{G}_{\mathrm{lc}}}$:
\begin{equation}
\small
{\mathbf{G}_{\mathrm{lc}}}
=\sigma_s\!\left(
\texttt{Conv}_{1\times1}
\left({\mathbf{O}_{\mathrm{shift}}}\odot \mathbf{O}_{\mathrm{att}} \right)
\right),
\end{equation}
where ${\mathbf{G}_{\mathrm{lc}}}$ is generated from the refinement  between the shift-based local response ${\mathbf{O}_{\mathrm{shift}}}$ and the local self-attention response $\mathbf{O}_{\mathrm{att}}$.

Consequently, the contour-guided fusion feature $\mathbf{F}_{\mathrm{CGF}}$ is obtained by applying the same modulation map to the CNN feature $\mathbf{F}_{\text{cnn}}$ and the VSSM feature $\mathbf{F}_{\text{vssm}}$, and the two recalibrated branch-specific features are concatenated and projected by a 1$\times$1  convolution:
\begin{equation}
\small
\mathbf{F}_{\mathrm{CGF}}
=
\texttt{Conv}_{1\times1}
\left(
[{\mathbf{G}_{\mathrm{lc}}}\odot\mathbf{F}_{\text{cnn}},{\mathbf{G}_{\mathrm{lc}}}\odot\mathbf{F}_{\text{vssm}}
]
\right).
\label{eq:cbgcm_output}
\end{equation}
The generated contour-guided feature $\mathbf{F}_{\mathrm{CGF}}$ provides stage-specific refinement representations for both encoder and decoder pathways, denoted as $E_F$ and $D_F$, respectively.
For each network stage, the generated refinement features are element-wise added to the corresponding encoder and decoder features at the same stage.
Specifically, the refinement feature associated with the $i$-th stage is injected into the $i$-th encoder and decoder of both the CNN and VSSM branches through residual connections. The refined feature representations are then propagated to the subsequent network stages for further feature extraction and cross-backbone interaction.

\subsection{Stable Gradient-Context Optimization}
Even with improved feature representations, fixed or empirically designed learning rate schemes are insufficient to ensure fast convergence and stable optimization in complex degradations  \cite{qi2024gradient}. To address this issue, we propose the GCO strategy that treats hyperparameter scheduling as an adaptive decision process. At the end of epoch $t$, the controller observes state $s_t$, selects action $a_t$, and maps it to the learning rate ${\mathrm{lr}_{t+1}}$ for the next epoch. After training epoch $t+1$, it receives reward $r_t$, observes the next state $s_{t+1}$, and updates the Q-table $\mathcal{Q}(s_t,a_t)$.

\subsubsection{Optimization State Construction}
Specifically, we denote the average training loss as ${\mathcal{L}_t}$ for the epoch $t$, the learning rate as ${\mathrm{lr}_t}$, and the mini-batch gradient norm as
$h_{t,b}=\|\nabla_{\theta}{\mathcal{L}_{t,b}}\|_2$, where $b$ indexes the mini-batch. We first compute the mean $\mu_t = \texttt{Mean}_b(h_{t,b})$ and the standard deviation $\sigma_t = \texttt{Std}_b(h_{t,b})$ of gradient norms within the current epoch and define a gradient-stability score $\kappa_{s}$:
\begin{equation}
\small
\kappa_s = \frac{\mu_t}{\sigma_t+\epsilon_0},
\label{eq:gradient_context}
\end{equation}
where $\epsilon_0$ denotes a small constant for numerical stability. A larger $\kappa_s$ indicates more consistent update magnitudes and a more stable optimization trajectory.

Together with $\kappa_s$,  we further compute the recent average learning rate:
\begin{equation}
\small
{\overline{\mathrm{lr}}_t}
=
\frac{1}{K_t}
\sum\nolimits_{i=t-K_t+1}^{t} {\mathrm{lr}_i} ,
\label{eq:recent_lr}
\end{equation}
where $K_t=\boldsymbol{\min}(K,t)$ denotes the effective history length at the epoch $t$. Then, two context variables $(\rho_t,{c^{\mathrm{lr}}_t})$ are defined to describe the loss progress and recent learning rate history:
\begin{equation}
\small
(\rho_t,{c^{\mathrm{lr}}_t})
=
\left(
\frac{{\mathcal{L}_t}}{\max_{1\le i\le t} {\mathcal{L}_i}+\epsilon_0},
\frac{\log {\overline{\mathrm{lr}}_t}-\log {\mathrm{lr}_{\min}}}
{\log {\mathrm{lr}_{\max}}-\log {\mathrm{lr}_{\min}}}
\right),
\label{eq:context_variables}
\end{equation}
where $\rho_t$ describes the relative loss progress, and ${c^{\mathrm{lr}}_t}$ encodes the recent learning rate regime in the normalized log space. Including ${c^{\mathrm{lr}}_t}$ makes the state partially aware of recent step-size decisions, which helps the controller distinguish whether the current optimization behavior occurs under an aggressive or conservative learning rate regime. $[{\mathrm{lr}_{\min}},{\mathrm{lr}_{\max}}]$ denotes the allowable learning rate range.

Finally, the three context variables $(\rho_t, \kappa_s, {c^{\mathrm{lr}}_t})$ are discretized into a finite state:
\begin{equation}
\small
\resizebox{0.88\hsize}{!}{$s_t=\texttt{Bin}(\rho_t;N_{\text{bin}})N_{\text{bin}}^2+\texttt{Bin}(\kappa_s;N_{\text{bin}})N_{\text{bin}}+\texttt{Bin}( {c^{\mathrm{lr}}_t};N_{\text{bin}})$},
\label{eq:state_index}
\end{equation}
where $\texttt{Bin}(\cdot;N_{\text{bin}})$ maps each scalar context variable into one of $N_{\text{bin}}$ discrete bins. Thus, the state space has the size of $N_{\text{bin}}^3$.

\subsubsection{Action-based Learning Rate Adjustment}
Given the state $s_t$, the agent selects an action $a_t\in\{0,\ldots, N_{\text{act}}-1\}$ from the Q-table using an ${\varepsilon_t}$-greedy policy, where $N_{\text{act}}$ denotes the total number of candidate learning rate actions:
\begin{equation}
\small
\resizebox{0.88\hsize}{!}{$a_t=
\begin{cases}
\texttt{Uniform}(0,N_{\text{act}}-1), & \text{with prob. } {\varepsilon_t},\\
\boldsymbol{\arg\max}_{a\in \{0,...,N_{\text{act}}-1\}} \mathcal{Q}(s_t,{a}), & \text{with prob. }1- {\varepsilon_t},
\end{cases}$}
\label{eq:action_lr_mapping}
\end{equation}
where the selected action is mapped to the learning rate for the next epoch in the logarithmic space, allowing the policy to cover the interval $[{\mathrm{lr}_{\min}}, {\mathrm{lr}_{\max}}]$ without additional gradient computation:
\begin{equation}
\small
{\mathrm{lr}_{t+1}}={\mathrm{lr}_{\min}}
\left(\frac{{\mathrm{lr}_{\max}}}{{\mathrm{lr}_{\min}}}\right)^{a_t/(N_{\text{act}}-1)}.
\end{equation}

\subsubsection{Reward Design and Online Q-table Update}
Let $\mathcal{W}_{t+1}=\{\texttt{max}(1,t-K+2),\ldots,t+1\}$ denote the recent temporal window after applying the action $a_t$. After the training epoch $t+1$ with ${\mathrm{lr}_{t+1}}$, the reward is defined as
\begin{equation}
\small
\resizebox{0.88\hsize}{!}{$r_t = {\lambda_{\mathrm{slope}}}\bigl(-\texttt{Slope}({\mathcal{L}_{\mathcal{W}_{t+1}}})\bigr)
+ {\lambda_{\mathrm{drop}}}({\mathcal{L}_t} - {\mathcal{L}_{t+1}}) - {\lambda_{\mathrm{stab}}}\texttt{Std}(\kappa_{\mathcal{W}_{t+1}})$},
\label{eq:reward_function}
\end{equation}
where $\texttt{Slope}({\mathcal{L}_{\mathcal{W}_{t+1}}})$ denotes the least-squares slope of the loss sequence within the recent window, ${\mathcal{L}_t}-{\mathcal{L}_{t+1}}$ measures short-term loss reduction after applying the selected learning rate, and $\texttt{Std}(\kappa_{\mathcal{W}_{t+1}})$  penalizes unstable gradient context. 
The slope term $\lambda_{\mathrm{slope}}$ encourages a sustained downward loss trend, the drop term $\lambda_{\mathrm{drop}}$ rewards immediate improvement, and the stable term $\lambda_{\mathrm{stab}}$ adjusts the gradient-context penalty.

The next state $s_{t+1}$ is obtained by discretizing $(\rho_{t+1},\kappa_{t+1}, {c^{\mathrm{lr}}_{t+1}})$ in the same way as Eq.~\eqref{eq:state_index}. The Q-table is initialized from scratch and updated online during network training, and it is not pre-trained on separate tasks. The Q-table update is defined as:
\begin{equation}
\resizebox{0.88\hsize}{!}{$
\begin{gathered}
\mathcal{Q}(s_t,a_t)
\leftarrow
\mathcal{Q}(s_t,a_t)
+
\alpha_q
\Big[
r_t
+
\gamma_q\boldsymbol{\max}_{a'}\mathcal{Q}(s_{t+1},a')
- \mathcal{Q}(s_t,a_t)
\Big],
\end{gathered}$}
\label{eq:q_learning_update}
\end{equation}
where $a'$ denotes a candidate action in the next state $s_{t+1}$,  $\max_{a'}\mathcal{Q}(s_{t+1},a')$ represents the estimated optimal future value, $\alpha_q$ denotes the Q-table update rate, and $\gamma_q$ represents the discount factor. The exploration rate is gradually decayed by
\begin{equation}
\small
\varepsilon_{t+1}
=
\texttt{max}(\zeta\varepsilon_t,\varepsilon_{\min}).
\label{eq:epsilon_decay}
\end{equation}
where $\varepsilon_{t}$ denotes the exploration rate at the epoch $t$, which controls the probability of randomly selecting a learning rate action in the ${\varepsilon_t}$-greedy policy. $\zeta\in(0,1)$ represents the decay factor that progressively reduces the exploration rate, $\varepsilon_{\min}$ denotes the minimum exploration rate, and the $\max(\cdot)$ operation prevents the exploration probability from becoming smaller than this lower bound.

\section{Experimental Validations}
\label{sec:experimental_results}
In the experiments, we evaluate three aspects of \ourMethod{}: 
(1) the robustness of MIS performance under heterogeneous degradations, 
(2) the effectiveness of explicit local-global decoupling with selective cross-backbone interaction, and 
(3) the impact of optimization-aware training on optimization  stability. 
Specifically, we have conducted comprehensive evaluations across various heterogeneous medical imaging datasets, diverse segmentation backbones, and extensive ablation settings, together with computational complexity analysis.
\begin{figure}[!t]
    \centering
    \includegraphics[width=0.98\linewidth, height=0.44\linewidth]{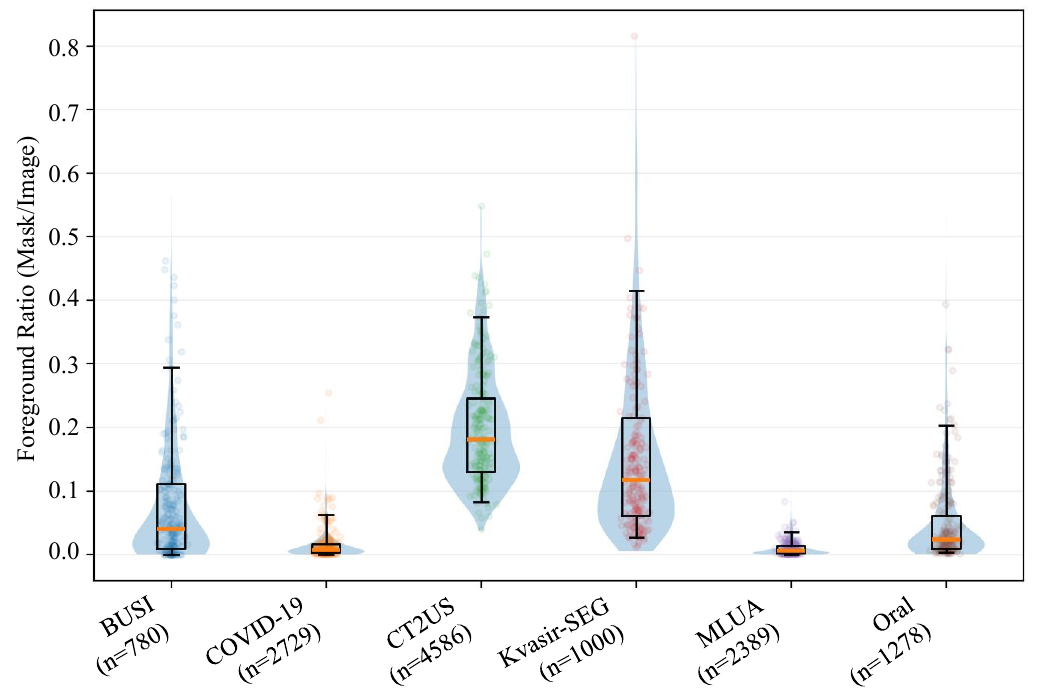}
    \caption{Foreground-ratio distributions across the six datasets. The evident cross-dataset shifts indicate substantial variation in target-scale statistics.}
    \label{fig5}
\end{figure}

\subsection{Experimental Protocols}
\subsubsection{Datasets}
We evaluate \ourMethod{} on six benchmark datasets spanning three imaging modalities and diverse anatomical structures, including \textit{optical RGB images} ({\textit{Oral} \cite{hu2026oral}} and \textit{Kvasir-SEG}\cite{jha2019kvasir}), \textit{radiological images} (\textit{COVID-19}\cite{jun2020covid} and \textit{MLUA}\cite{wang2023multi}), and \textit{ultrasound images} (\textit{BUSI}\cite{al2020dataset} and \textit{CT2US}\cite{song2022ct2us}).

\begin{itemize}[leftmargin=*]
\item \textit{Oral}: An in-house dataset with 1,278 expert-annotated oral mucosal {RGB images}, characterized by small, scattered lesions and strong specular highlights.
\item \textit{Kvasir-SEG}: A polyp segmentation dataset with 1,000 endoscopic {RGB images}, featuring large variations in lesion scale, shape, and illumination.
\item \textit{COVID-19}: A collection of 2,729 annotated {CT images} aggregated from three public datasets, exhibiting low contrast, blurred boundaries, and imaging artifacts.
\item \textit{MLUA}: A dental panoramic X-ray image dataset for caries segmentation containing 2,389 images, characterized by subtle contrast differences and complex lesion–background boundaries.
\item \textit{CT2US}: A dataset with 4,586 kidney segmentation {ultrasound image samples}, where synthesized ultrasound-like appearance introduces complex texture variations.
\item \textit{BUSI}: A breast {ultrasound image} dataset containing 780 images from 600 patients, with severe speckle noise and intensity inhomogeneity.
\end{itemize}

\begin{table*}[!t]
\centering
\scriptsize
\setstretch{0.88}
\setlength{\tabcolsep}{3.6pt}
\renewcommand{\arraystretch}{1.06}
\caption{Quantitative comparison in terms of mIoU and Dice on the \textit{Oral}, \textit{Kvasir-SEG}, \textit{COVID-19}, \textit{MLUA}, \textit{CT2US}, and \textit{BUSI} datasets. Comparison methods are grouped into \textit{Reference Baselines}, \textit{Single-stream Frameworks}, and \textit{Hybrid-stream Frameworks}. The best and second-best results in each column are highlighted in \best{red} and \second{blue}, respectively.}
\label{tab:quantitative_comparison}
\resizebox{\textwidth}{!}{
\begin{tabular}{p{1.35cm}lcccccccccccc}
\toprule
\multirow{3}{*}{Group} & \multirow{3}{*}{Method} 
& \multicolumn{4}{c}{\textit{Optical RGB Dataset}} 
& \multicolumn{4}{c}{\textit{Radiological Dataset}} 
& \multicolumn{4}{c}{\textit{Ultrasound Dataset}} \\
\cmidrule(lr){3-6}\cmidrule(lr){7-10}\cmidrule(lr){11-14}
&& \multicolumn{2}{c}{\textit{Oral}}
& \multicolumn{2}{c}{\textit{Kvasir-SEG}}
& \multicolumn{2}{c}{\textit{COVID-19}}
& \multicolumn{2}{c}{\textit{MLUA}}
& \multicolumn{2}{c}{\textit{CT2US}}
& \multicolumn{2}{c}{\textit{BUSI}} \\
\cmidrule(lr){3-4}\cmidrule(lr){5-6}\cmidrule(lr){7-8}\cmidrule(lr){9-10}\cmidrule(lr){11-12}\cmidrule(lr){13-14}
& 
& mIoU$\uparrow$ & Dice$\uparrow$
& mIoU$\uparrow$ & Dice$\uparrow$
& mIoU$\uparrow$ & Dice$\uparrow$
& mIoU$\uparrow$ & Dice$\uparrow$
& mIoU$\uparrow$ & Dice$\uparrow$
& mIoU$\uparrow$ & Dice$\uparrow$ \\
\midrule

\multirow{5}{*}{\makecell[c]{\textit{Reference}\\\textit{Baselines}}}
& \textit{U$^2$-Net}\cite{qin2020u2}
& 82.3$\pm$3.1 & 90.3$\pm$1.7
& 78.5$\pm$3.8 & 88.0$\pm$3.2
& \second{79.4$\pm$3.8} & \second{88.5$\pm$3.3}
& 59.5$\pm$5.8 & 68.5$\pm$5.2
& 95.3$\pm$4.0 & 97.6$\pm$2.7
& 61.9$\pm$4.1 & 76.4$\pm$3.6 \\

& \textit{DeepLabV3+}\cite{chen2018encoder}
& 81.6$\pm$2.3 & 89.8$\pm$1.4
& 50.1$\pm$6.7 & 66.8$\pm$6.1
& 78.3$\pm$4.6 & 87.8$\pm$4.3
& 59.1$\pm$4.8 & 74.3$\pm$4.6
& 95.1$\pm$3.9 & 97.5$\pm$1.7
& 61.1$\pm$5.6 & 75.8$\pm$5.6 \\

& \textit{SFM}\cite{chen2025spatial}
& 70.0$\pm$3.0 & 82.4$\pm$2.4
& 68.8$\pm$5.1 & 81.5$\pm$4.7
& 75.4$\pm$5.3 & 85.9$\pm$3.2
& 45.6$\pm$6.2 & 62.6$\pm$5.6
& 94.4$\pm$2.1 & 97.1$\pm$1.4
& 58.7$\pm$3.9 & 74.0$\pm$3.6 \\

& \textit{TherNet}\cite{chen2025thernet}
& 82.5$\pm$2.9 & 90.4$\pm$1.9
& 74.4$\pm$3.8 & 85.3$\pm$2.9
& 78.3$\pm$2.9 & 87.8$\pm$2.5
& 46.9$\pm$6.7 & 63.8$\pm$6.7
& 95.0$\pm$4.1 & 97.4$\pm$2.2
& 50.9$\pm$6.8 & 67.4$\pm$6.4 \\

& \textit{MWCNN}\cite{liu2018multi}
& 78.6$\pm$3.5 & 88.0$\pm$2.6
& 70.8$\pm$4.6 & 82.9$\pm$3.6
& 71.4$\pm$4.1 & 83.3$\pm$3.4
& 40.8$\pm$7.2 & 58.0$\pm$6.8
& 94.8$\pm$3.8 & 97.3$\pm$1.9
& 40.5$\pm$7.6 & 57.6$\pm$7.3 \\

&\textit{MedSAM}\cite{ma2024segment}
& 79.4$\pm$4.5 & 88.2$\pm$2.9
& 77.6$\pm$3.1 & 81.9$\pm$2.6
& 75.7$\pm$5.2 & 86.0$\pm$4.6
& 62.5$\pm$5.5 & 74.8$\pm$6.3
& 94.0$\pm$3.0 & 97.2$\pm$1.9
& \second{67.1$\pm$3.8} & 73.9$\pm$3.6 \\

\midrule

\multirow{20}{*}{\makecell[c]{\textit{Single-stream}\\\textit{Frameworks}}}

& \textit{BCDU-Net}\cite{azad2019bi}
& 85.8$\pm$2.6 & 92.3$\pm$1.6
& 72.6$\pm$4.7 & 84.1$\pm$3.8
& 78.8$\pm$3.0 & 88.1$\pm$2.7
& 58.2$\pm$5.7 & 73.6$\pm$4.4
& 94.4$\pm$3.3 & 97.1$\pm$2.1
& 58.1$\pm$5.3 & 73.5$\pm$4.8 \\

& \textit{MTG-Net}\cite{chen2025neovascularization}
& 73.1$\pm$7.4 & 81.2$\pm$4.0
& 68.5$\pm$8.5 & 74.4$\pm$5.9
& 72.4$\pm$6.7 & 76.6$\pm$4.8
& 59.6$\pm$6.9 & 74.7$\pm$6.1
& 74.1$\pm$6.0 & 85.1$\pm$2.9
& 48.2$\pm$7.9 & 55.3$\pm$6.4 \\

& \textit{FAT-Net}\cite{wu2022fat}
& 83.3$\pm$3.5 & 90.9$\pm$2.2
& 78.0$\pm$3.6 & 87.6$\pm$3.6
& 78.1$\pm$4.6 & 87.7$\pm$3.6
& 60.4$\pm$4.9 & 75.3$\pm$5.0
& 95.3$\pm$3.5 & 97.6$\pm$2.1
& 58.6$\pm$4.6 & 73.9$\pm$4.2 \\

& \textit{MISSFormer}\cite{huang2022missformer}
& 67.5$\pm$2.5 & 80.6$\pm$1.6
& 75.8$\pm$2.3 & 86.2$\pm$2.3
& 76.4$\pm$3.1 & 86.6$\pm$2.8
& 44.9$\pm$6.8 & 62.0$\pm$6.3
& 94.3$\pm$3.7 & 97.0$\pm$0.9
& 55.9$\pm$4.1 & 71.7$\pm$3.6 \\

& \textit{LeViT-UNet}\cite{xu2023levit}
& \second{86.9$\pm$2.4} & \second{93.0$\pm$1.2}
& 76.5$\pm$3.0 & 86.7$\pm$2.5
& 78.6$\pm$3.6 & 88.0$\pm$2.9
& 59.4$\pm$6.3 & 74.5$\pm$5.5
& 95.2$\pm$3.2 & 97.5$\pm$1.8
& 60.9$\pm$5.3 & 75.7$\pm$4.1 \\

& \textit{Swin-UNet}\cite{cao2022swin}
& 62.1$\pm$5.1 & 69.2$\pm$3.8
& 69.9$\pm$7.0 & 77.1$\pm$6.2
& 72.1$\pm$5.5 & 83.8$\pm$5.8
& 43.2$\pm$6.6 & 53.3$\pm$5.7
& 92.1$\pm$5.5 & 95.8$\pm$2.5
& 46.3$\pm$8.5 & 51.7$\pm$8.3 \\

& \textit{CyCoSeg}\cite{medley2021cycoseg}
& 66.5$\pm$4.5 & 79.9$\pm$3.0
& 68.6$\pm$4.2 & 81.3$\pm$3.1
& 75.2$\pm$4.7 & 85.9$\pm$4.1
& 47.1$\pm$7.1 & 64.0$\pm$5.8
& 94.1$\pm$3.4 & 96.9$\pm$1.6
& 57.9$\pm$5.9 & 73.3$\pm$4.7 \\

& \textit{Mamba-UNet}\cite{wang2024mamba}
& 80.9$\pm$3.0 & 89.4$\pm$1.8
& 78.1$\pm$4.5 & 87.7$\pm$4.2
& 76.0$\pm$5.5 & 86.3$\pm$3.1
& 58.4$\pm$5.6 & 73.7$\pm$5.2
& 94.7$\pm$3.8 & 97.3$\pm$2.0
& 61.6$\pm$5.2 & 76.2$\pm$5.1 \\

& \textit{UltraLight VM}\cite{wu2025ultralight}
& 79.4$\pm$3.7 & 88.5$\pm$2.0
& 79.6$\pm$3.9 & 88.6$\pm$3.7
& 78.9$\pm$3.9 & 88.2$\pm$5.3
& 60.4$\pm$4.8 & 75.3$\pm$4.4
& 94.6$\pm$4.5 & 97.2$\pm$2.5
& 54.8$\pm$4.2 & 70.8$\pm$3.7 \\

& \textit{LKM-UNet}\cite{wang2024lkm}
& 86.4$\pm$2.2 & 92.7$\pm$1.5
& 79.5$\pm$3.3 & 88.6$\pm$2.8
& 79.3$\pm$3.6 & 88.4$\pm$3.4
& 59.5$\pm$6.4 & 74.6$\pm$5.5
& \second{95.7$\pm$2.9} & \second{97.8$\pm$2.2}
& 60.2$\pm$6.6 & 75.1$\pm$6.3 \\

& \textit{H-vmuNet}\cite{wu2025h}
& 83.1$\pm$2.5 & 90.8$\pm$1.8
& 72.7$\pm$4.7 & 84.2$\pm$3.6
& 77.5$\pm$3.2 & 87.3$\pm$2.5
& 52.5$\pm$5.6 & 68.9$\pm$4.7
& 95.3$\pm$3.3 & 97.6$\pm$2.1
& 63.1$\pm$5.8 & 77.4$\pm$5.3 \\

& \textit{Mamba-Sea}\cite{cheng2025mamba}
& 75.7$\pm$6.4 & 83.1$\pm$4.0
& 75.4$\pm$7.3 & 82.3$\pm$5.7
& 63.4$\pm$5.4 & 73.6$\pm$6.4
& 55.5$\pm$5.9 & 71.4$\pm$5.3
& 94.9$\pm$1.8 & 97.3$\pm$1.3
& 54.2$\pm$8.5 & 61.0$\pm$8.8 \\

& \textit{FusionNet}\cite{quan2021fusionnet}
& 77.3$\pm$4.3 & 87.2$\pm$2.4
& 78.4$\pm$2.9 & 88.1$\pm$2.4
& 77.1$\pm$3.6 & 87.1$\pm$4.1
& 60.0$\pm$6.0 & 75.0$\pm$4.8
& 95.2$\pm$1.5 & 97.5$\pm$1.0
& 61.6$\pm$4.6 & 76.3$\pm$4.2 \\

& \textit{MBP-SSNet}\cite{wang2024multi}
& 81.5$\pm$2.1 & 89.8$\pm$2.1
& 71.1$\pm$3.1 & 83.1$\pm$2.6
& 75.4$\pm$3.7 & 85.9$\pm$3.0
& 63.3$\pm$5.8 & 77.5$\pm$5.0
& 92.1$\pm$3.5 & 95.9$\pm$2.5
& 56.2$\pm$4.8 & 72.0$\pm$4.5 \\

& \textit{EAG-SAE-ESTF}\cite{zhang2025eag}
& 74.9$\pm$2.5 & 85.6$\pm$1.8
& 79.0$\pm$2.5 & 88.3$\pm$2.2
& 77.7$\pm$2.8 & 87.4$\pm$2.6
& 61.1$\pm$4.9 & 75.9$\pm$4.2
& 95.4$\pm$2.8 & 97.6$\pm$1.1
& 63.7$\pm$4.0 & 77.8$\pm$3.6 \\

& \textit{GobletNet}\cite{zhou2024gobletnet}
& 85.3$\pm$2.2 & 92.0$\pm$1.3
& \second{80.0$\pm$2.8} & \second{88.9$\pm$2.2}
& 79.0$\pm$2.8 & 88.2$\pm$2.3
& 55.1$\pm$4.9 & 71.1$\pm$4.5
& 95.0$\pm$2.5 & 97.4$\pm$1.1
& 65.3$\pm$3.8 & \second{79.0$\pm$4.2} \\

& \textit{MADGNet}\cite{nam2024modality}
& 82.4$\pm$2.6 & 90.3$\pm$1.4
& 77.6$\pm$3.3 & 87.4$\pm$2.9
& 78.9$\pm$4.3 & 88.2$\pm$4.6
& 61.3$\pm$5.1 & 76.0$\pm$3.6
& 95.1$\pm$3.0 & 97.5$\pm$2.1
& 60.8$\pm$3.7 & 75.6$\pm$3.1 \\

& \textit{FFTMed}\cite{pham2025fftmed}
& 51.9$\pm$4.9 & 68.4$\pm$3.0
& 51.4$\pm$5.5 & 67.9$\pm$6.1
& 69.9$\pm$5.9 & 82.3$\pm$4.5
& 59.1$\pm$5.7 & 65.1$\pm$5.3
& 93.6$\pm$1.9 & 96.7$\pm$1.9
& 54.6$\pm$4.3 & 70.6$\pm$4.7 \\

& \textit{CSWin-UNet}\cite{liu2025cswin}
& 66.4$\pm$5.3 & 70.2$\pm$2.0
& 70.9$\pm$7.3 & 78.6$\pm$5.7
& 59.9$\pm$3.5 & 71.2$\pm$5.4
& 55.6$\pm$5.4 & 71.4$\pm$4.6
& 80.5$\pm$4.2 & 85.4$\pm$3.7
& 58.8$\pm$3.9 & 74.1$\pm$3.5 \\

&\textit{SAM2-UNet}\cite{xiong2026sam2}
& 67.8$\pm$4.4 & 78.5$\pm$3.2
& 79.4$\pm$2.8 & 87.9$\pm$2.4
& 68.1$\pm$4.6 & 79.6$\pm$3.4
& 47.9$\pm$6.3 & 50.1$\pm$7.6
& 95.6$\pm$1.4 & 97.7$\pm$1.3
& 64.8$\pm$4.5 & 71.5$\pm$3.2 \\
\midrule

\multirow{5}{*}{\makecell[c]{\textit{Hybrid-stream}\\\textit{Frameworks}}}
& \textit{MedT}\cite{valanarasu2021medical}
& 68.4$\pm$5.4 & 75.2$\pm$1.9
& 67.9$\pm$4.2 & 80.8$\pm$3.9
& 76.8$\pm$2.8 & 86.9$\pm$2.5
& 54.3$\pm$6.6 & 61.0$\pm$4.9
& 69.3$\pm$6.6 & 81.8$\pm$4.3
& 50.2$\pm$6.6 & 66.8$\pm$6.0 \\

& \textit{TransFuse}\cite{zhang2021transfuse}
& 83.1$\pm$2.4 & 90.7$\pm$1.6
& 78.1$\pm$2.6 & 87.7$\pm$2.4
& 78.2$\pm$4.6 & 87.8$\pm$3.7
& 60.3$\pm$4.8 & 75.2$\pm$4.2
& 95.1$\pm$3.1 & 97.4$\pm$2.2
& 62.7$\pm$5.3 & 77.1$\pm$5.1 \\

& \textit{H2Former}\cite{he2023h2former}
& 80.1$\pm$2.7 & 88.9$\pm$1.7
& 72.3$\pm$3.6 & 83.9$\pm$2.7
& 78.2$\pm$2.9 & 87.7$\pm$2.6
& 59.7$\pm$6.7 & 74.7$\pm$5.3
& 95.3$\pm$2.6 & 97.6$\pm$2.1
& 57.3$\pm$5.7 & 72.9$\pm$5.9 \\

& \textit{MambaEviScrib}\cite{han2026mambaeviscrib}
& 76.8$\pm$2.7 & 86.9$\pm$1.4
& 78.8$\pm$3.5 & 88.2$\pm$3.1
& 77.7$\pm$3.0 & 87.5$\pm$2.4
& 59.7$\pm$5.6 & 74.8$\pm$5.1
& 95.2$\pm$3.3 & 97.5$\pm$1.6
& 60.8$\pm$5.0 & 75.6$\pm$4.3 \\

& \textit{XNet}\cite{zhou2023xnet}
& 85.0$\pm$2.7 & 91.9$\pm$2.0
& 79.0$\pm$3.2 & 88.2$\pm$3.0
& 78.9$\pm$3.4 & 88.2$\pm$3.7
& \second{64.1$\pm$4.8} & \second{78.1$\pm$4.1}
& 95.3$\pm$3.1 & 97.6$\pm$1.5
& 63.1$\pm$6.5 & 77.4$\pm$5.8 \\

& \textit{DS2F}\cite{qiu2023rethinking}
& 78.9$\pm$2.1 & 88.2$\pm$1.5
& 75.8$\pm$2.9 & 86.2$\pm$2.6
& 78.1$\pm$3.3 & 87.7$\pm$2.5
& 50.9$\pm$6.3 & 67.5$\pm$5.3
& 94.5$\pm$2.3 & 97.1$\pm$1.7
& 60.4$\pm$4.6 & 75.3$\pm$4.2 \\
\midrule

& \textbf{Ours}
& \best{91.9$\pm$1.9} & \best{96.8$\pm$1.1}
& \best{83.4$\pm$2.3} & \best{90.9$\pm$2.1}
& \best{82.6$\pm$2.5} & \best{90.4$\pm$2.1}
& \best{67.9$\pm$4.5} & \best{80.9$\pm$3.1}
& \best{96.2$\pm$1.2} & \best{98.0$\pm$0.9}
& \best{69.3$\pm$3.5} & \best{81.9$\pm$3.1} \\

\bottomrule
\end{tabular}
}
\end{table*}

The six datasets exhibit substantial differences in foreground scale and object scale, resulting in significant changes in distribution across datasets. The heterogeneity of these datasets can effectively validate the robustness of subsequent comparison approaches. 
As shown in \textcolor{blue}{Fig.} \ref{fig5}, \textit{CT2US} generally contains larger foreground regions, whereas \textit{COVID-19}, \textit{MLUA}, and \textit{Oral} are dominated by small or sparse targets. In contrast, \textit{BUSI} and \textit{Kvasir-SEG} present broader and more heavy-tailed distributions.

\subsubsection{Evaluation Metrics}
We adopt Dice similarity coefficient (DSC, \textit{Dice}) \cite{xu2024deep}, mean intersection-over-union (\textit{mIoU}) \cite{eelbode2020optimization}, and \textit{Sensitivity} \cite{wang2022boundary} to measure region overlap and foreground detection capability, which have been widely used in existing MIS frameworks \cite{rueckert2026comparative,maier2024metrics,ma2024segment}. In addition, \textit{Betti Error} (BE) \cite{li2025topology} is widely used to evaluate topological consistency by measuring the discrepancy between predicted and ground-truth Betti numbers. For multiclass settings, Dice and mIoU are macro-averaged over foreground classes excluding the lesion-background, and for binary settings, all metrics are computed on the foreground class.

\subsubsection{Comparison Methods}
We compare \ourMethod{} with a broad set of representative MIS approaches, following the taxonomy established in Section~\ref{sec:related_work}. Specifically, all competing methods are grouped into three categories: (1) \textit{Reference Baselines}, (2) \textit{Single-stream Frameworks}, and (3) \textit{Hybrid-stream Frameworks}. 
\begin{itemize}[leftmargin=*]
\item \textit{Reference Baselines.}
We adopt six widely-used segmentation backbones and classical architectures, including \textit{U$^2$-Net}~\cite{qin2020u2}, \textit{DeepLabV3+}~\cite{chen2018encoder}, \textit{SFM}~\cite{chen2025spatial}, \textit{TherNet}~\cite{chen2025thernet},  \textit{MWCNN}~\cite{liu2018multi}, and MedSAM~\cite{ma2024segment}. These methods provide standardized baselines for evaluating general segmentation capability and robustness under varying imaging conditions.

\item \textit{Single-stream Frameworks.}
We consider representative CNN-based, Transformer-based, VSSM-based, and frequency-enhanced single-stream models, such as SAM2-UNet~\cite{xiong2026sam2}, \textit{BCDU-Net}~\cite{azad2019bi}, MTG-Net \cite{chen2025neovascularization}, \textit{MISSFormer}~\cite{huang2022missformer}, \textit{Swin-UNet}~\cite{cao2022swin}, \textit{Mamba-UNet}~\cite{wang2024mamba}, \textit{LKM-UNet}~\cite{wang2024lkm}, \textit{GobletNet}~\cite{zhou2024gobletnet}, \textit{MADGNet}~\cite{nam2024modality}, and \textit{FFTMed}~\cite{pham2025fftmed}. These methods represent diverse strategies for enhancing local detail, global structure, and domain-based  representations within a single network pathway.

\item \textit{Hybrid-stream Frameworks.}
We further compare with representative hybrid architectures that combine heterogeneous representations, including \textit{MedT}~\cite{valanarasu2021medical}, \textit{TransFuse}~\cite{zhang2021transfuse}, \textit{H2Former}~\cite{he2023h2former}, \textit{MambaEviScrib}~\cite{han2026mambaeviscrib}, \textit{XNet}~\cite{zhou2023xnet}, and \textit{DS2F}~\cite{qiu2023rethinking}. These methods incorporate multi-branch designs to exploit complementary local-global or multi-domain information, providing strong baselines for evaluating the effectiveness of cross-stream interaction and fusion strategies.
\end{itemize}

\begin{figure*}[!t]
\centering
\includegraphics[width=0.92\textwidth,height=0.8\textwidth]{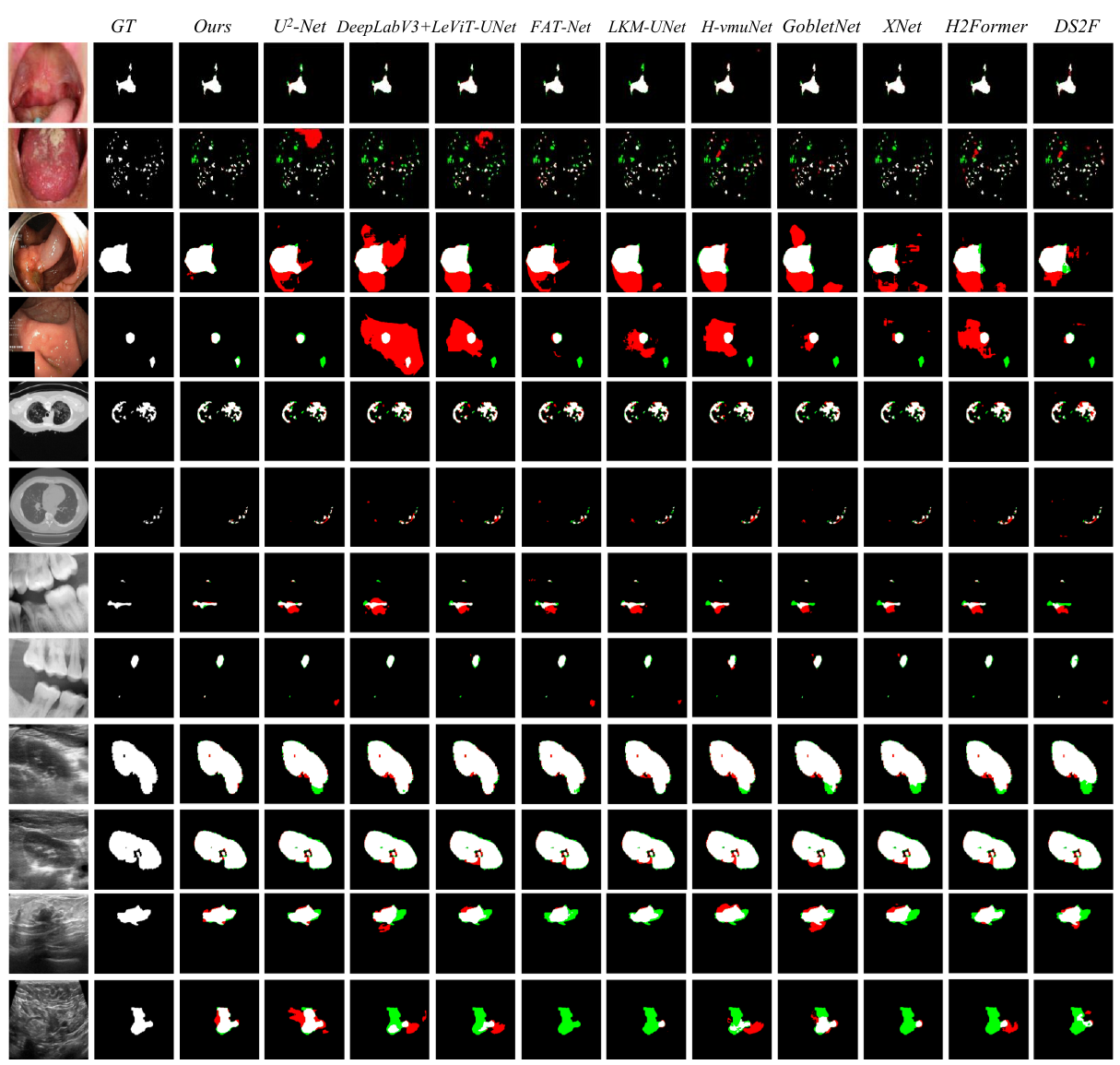}
\caption{Qualitative comparison across different modalities of medical imaging data. For representative coverage, the competitive methods consist of  \textit{Reference Baselines}, \textit{Single-stream Frameworks}, and \textit{Hybrid-stream Frameworks}. White indicates correct overlap, red indicates false positives, and green indicates false negatives.}
\label{fig6}
\end{figure*}

\subsubsection{Implementation Details}
\ourMethod{} is implemented in \textit{PyTorch} and trained on an Intel i9-10980XE CPU and an NVIDIA GeForce RTX4090 GPU. Adam is  used as the base optimizer, with an initial learning rate of $1 \times 10^{-3}$ and a weight decay of $1 \times 10^{-2}$. For all experiments, images are resized to $224 \times 224$ and each input is converted to a three-channel representation by channel replication when necessary and normalized to $[0,1]$. We adopt  five-fold cross-validation for all datasets. In each round, one fold is used for testing and the remaining four folds are used for training. The related network is optimized using the binary cross-entropy with logits loss (\textit{i.e.}, \textit{BCEWithLogits} loss), while the learning rate schedule is adaptively controlled by our GCO module. Finally, training is conducted for 500 epochs with a batch size of 4.

For all experiments, GCO uses fixed hyperparameters across datasets and folds. The state and action spaces are discretized with $N_{\text{bin}}$=5 levels and $N_{\text{act}}$=6 logarithmically spaced learning rates in $[lr_{\min},lr_{\max}]$=$[10^{-8},10^{-3}]$, yielding $N_{\text{bin}}^3$=125 states. The optimization window is $K$=5, with $\alpha_q$=0.01 and $\gamma_q$=0.95. Exploration follows $\varepsilon_1$=1.0, $\zeta$=0.9995, and $\varepsilon_{\min}$=0.1. The reward weights are $\lambda_{\mathrm{slope}}$=100, $\lambda_{\text{drop}}$=10, and $\lambda_{\text{stab}}$=0.5, with $\epsilon_0$=$10^{-8}$ for numerical stability.

In addition, all comparison methods are evaluated on six benchmark datasets (\textit{i.e.}, \textit{Oral}, \textit{Kvasir-SEG}, \textit{COVID-19}, \textit{MLUA}, \textit{CT2US}, and \textit{BUSI}) using mIoU and Dice as evaluation metrics. To ensure fairness, we strictly follow the official implementations and training protocols when available, and otherwise adopt unified training settings, including identical data splits, input resolutions, and optimization schedules. This setup enables a consistent and comprehensive comparison across different methodological paradigms.

\subsection{Comparison Performance}
\subsubsection{Quantitative Comparison}
\textcolor{blue}{Table}~\ref{tab:quantitative_comparison} reports the quantitative results on six MIS dataset benchmarks. For clarity, competing methods are grouped into \textit{Reference Baselines}, \textit{Single-stream Frameworks},  \textit{Hybrid-stream Frameworks}, and our \ourMethod{}. As seen, \ourMethod{} consistently achieves the best performance across all datasets, yielding the highest mIoU and Dice values. This uniform dominance across heterogeneous imaging modalities indicates strong robustness to variations in degradation patterns, target scale, and anatomical complexity.

Compared with the highest-competing method in \textcolor{blue}{Table}~\ref{tab:quantitative_comparison} on each dataset, \ourMethod{} improves mIoU/Dice by 5.0/3.8 on \textit{Oral}, 3.4/2.0 on \textit{Kvasir-SEG}, 3.2/1.9 on \textit{COVID-19}, 3.8/2.8 on \textit{MLUA},  0.5/0.2 on \textit{CT2US}, and 2.2/2.9 on \textit{BUSI}. Notably, the gains are more significant on datasets with small or sparse targets, ambiguous boundaries, and severe degradations (\textit{e.g.}, \textit{Oral}, \textit{BUSI}, and \textit{MLUA}), while remaining consistent on \textit{CT2US}. In addition, \ourMethod{} often achieves lower or comparable standard deviations, suggesting greater robustness rather than optimization for specific data splits.

\subsubsection{Qualitative Comparison}
\textcolor{blue}{Fig.}~\ref{fig6} presents qualitative comparisons across different image modalities. To ensure representative coverage, we  select two \textit{Reference Baselines} (\textit{U$^2$-Net} and \textit{DeepLabV3+}), five \textit{Single-stream Frameworks} (\textit{LeViT-UNet}, \textit{FAT-Net}, \textit{LKM-UNet}, \textit{H-vmuNet}, and \textit{GobletNet}), and three \textit{Hybrid-stream Frameworks} (\textit{XNet}, \textit{H2Former}, and \textit{DS2F}). Overall, \ourMethod{} produces predictions that are more consistent with ground truth (GT), with cleaner boundaries, fewer false positives, and reduced missed regions.

For optical RGB images, MIS methods are often affected by illumination variations and specular reflections, whereas \ourMethod{} suppresses such degradations and improves lesion-background separation. For radiological images, it maintains both global coherence and local detail, reducing fragmentation and omission. For ultrasound images, it produces more continuous boundaries under speckle noise. Benefiting from dual-domain refinement, local-global decoupling, and contour-guided selective interaction, \ourMethod{} better preserves both regional coherence and fine lesion delineation across various modalities.
\begin{table}[!t]
\centering
\scriptsize
\setlength{\tabcolsep}{2.8pt}
\renewcommand{\arraystretch}{1.08}
\caption{Ablation study on the \textit{Oral} and \textit{COVID-19} datasets. Higher values indicate better performance for \textit{mIoU}, \textit{Dice}, and \textit{Sensitivity} (`Sens.'), while lower values indicate better performance for \textit{Betti Error} (BE).}
\resizebox{\columnwidth}{!}{
\begin{tabular}{c|ccccc|cccc}
\toprule
\multicolumn{10}{c}{\textbf{(a) Oral}} \\
\midrule
Order & CNN & VSSM & CGF & DDI & GCO & mIoU$\uparrow$ & Dice$\uparrow$ & Sens.$\uparrow$ & BE$\downarrow$ \\
\midrule
\circled{1} & $\checkmark$ &  &  &  &  & 86.0 & 92.4 & 92.3 & 13.0 \\
\circled{2} & $\checkmark$ & $\checkmark$ &  &  &  & 87.7 & 93.4 & 91.1 & 10.3 \\
\circled{3} & $\checkmark$ & $\checkmark$ & $\checkmark$ &  &  & 88.5 & 93.9 & 93.3 & 7.9 \\
\circled{4} & $\checkmark$ & $\checkmark$ & $\checkmark$ & $\checkmark$ &  & 90.1 & 94.8 & 93.2 & 5.3 \\
\rowcolor{black!16}
\circled{5} & $\checkmark$ & $\checkmark$ & $\checkmark$ & $\checkmark$ & $\checkmark$ & \textbf{91.9} & \textbf{96.8} & \textbf{95.5} & \textbf{3.4} \\
\bottomrule
\end{tabular}
}

\vspace{2pt}

\resizebox{\columnwidth}{!}{
\begin{tabular}{c|ccccc|cccc}
\multicolumn{10}{c}{\textbf{(b) COVID-19}} \\
\midrule
Order & CNN & VSSM & CGF & DDI & GCO & mIoU$\uparrow$ & Dice$\uparrow$ & Sens.$\uparrow$ & BE$\downarrow$ \\
\midrule
\circled{1} & $\checkmark$ &  &  &  &  & 79.2 & 88.4 & 88.0 & 15.8 \\
\circled{2} & $\checkmark$ & $\checkmark$ &  &  &  & 79.7 & 88.7 & 88.6 & 11.4 \\
\circled{3} & $\checkmark$ & $\checkmark$ & $\checkmark$ &  &  & 80.2 & 88.9 & 89.2 & 8.8 \\
\circled{4} & $\checkmark$ & $\checkmark$ & $\checkmark$ & $\checkmark$ &  & 81.9 & 90.1 & 89.7 & 5.8 \\
\rowcolor{black!16}
\circled{5} & $\checkmark$ & $\checkmark$ & $\checkmark$ & $\checkmark$ & $\checkmark$ & \textbf{82.6} & \textbf{90.4} & \textbf{89.9} & \textbf{4.0} \\
\bottomrule
\end{tabular}
}
\label{tab:ablation_singlecol}
\end{table}

\subsection{Ablation Studies}
\subsubsection{Quantitative Analysis}
\textcolor{blue}{Table}~\ref{tab:ablation_singlecol} presents  component-wise ablation results on \textit{Oral} and \textit{COVID-19}. The ablation pipeline follows the construction of \ourMethod{}, starting from a CNN baseline and progressively adding the VSSM branch, CGF, DDI, and GCO. As the model is extended from setting~\circled{1} to  setting~\circled{5}, \textit{mIoU} and \textit{Dice} results consistently improve, along with significant reductions in BE. This indicates that the proposed components improve not only overlap accuracy but also structural and topological consistency.

Introducing the VSSM branch reduces the BE results from 13.0 to 10.3  on \textit{Oral} and from 15.8 to 11.4 on\textit{ COVID-19}, showing that long-range dependency modeling improves global structural coherence. However, it slightly reduces sensitivity from 92.3 to 91.1  on \textit{Oral}, suggesting that global modeling alone is insufficient for stable detection of subtle lesions.

After incorporating CGF, BE further decreases to 7.9, while sensitivity recovers to 93.3 on \textit{Oral}, validating its role in coordinating local and global features. Adding DDI yields substantial gains, especially on \textit{COVID-19}, where Dice increases from 88.9 to 90.1 and BE decreases from 8.8 to 5.8. This is consistent with its role in suppressing degradation-sensitive responses on composite distortions. Finally, GCO further reduces BE from 5.3 to 3.4 on \textit{Oral}, and  it also improves the sensitivity from 93.2 to 95.5. These results indicate that GCO mainly contributes by alleviating under-convergence and training oscillation, thereby improving recall and structural stability.

\begin{figure}[!t]
\centering
\includegraphics[width=0.98\linewidth]{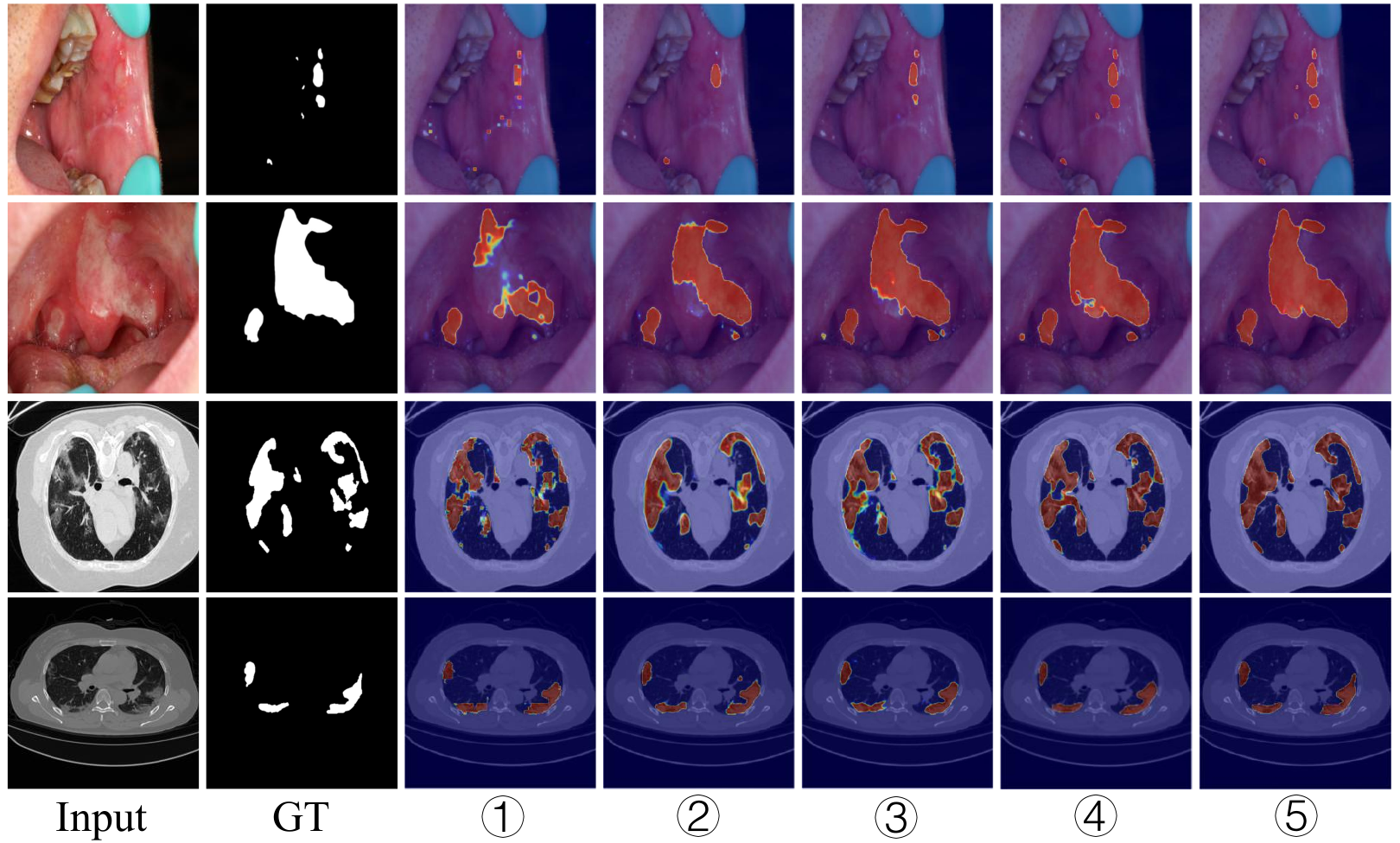}
\caption{Qualitative ablation results on representative samples from  \textit{Oral} and \textit{COVID-19} datasets. The predictions show progressively improved completeness and structural consistency.}
\label{fig7}
\end{figure}

\subsubsection{Qualitative Analysis}
\textcolor{blue}{Fig.}~\ref{fig7} presents qualitative results that are consistent with the quantitative trends in \textcolor{blue}{Table}~\ref{tab:ablation_singlecol}. In the CNN-only setting~\circled{1}, the predictions exhibit scattered activations and fragmented lesion regions, particularly in the oral lesion cases and the CT case. Incorporating the VSSM branch in setting~\circled{2} improves spatial coherence, although boundary delineation remains coarse and some lesion regions are still misaligned with the GT results.

With CGF in the setting~\circled{3}, lesion-background interfaces become more stable, and spurious connectivity or hole-like errors are reduced, indicating more effective local-global decoupling.
Adding DDI in setting~\circled{4} suppresses spurious activations and produces cleaner responses around lesion boundaries.  The full \ourMethod{} in setting~\circled{5} yields the most balanced predictions, preserving lesion structures while reducing lesion-background interference, resulting in more compact and accurate segmentation results across both oral and CT cases.

\subsection{Effectiveness Analysis} 
Although the ablation study quantifies the contribution of each component, it does not fully explain the underlying mechanisms. Therefore, we further analyze the effectiveness of the key modules (\textit{i.e.}, DDI, CGF, and GCO) from the perspectives of input-feature separability, cross-backbone interaction, and optimization dynamics. 
\begin{figure}[!t]
\centering
\includegraphics[width=0.98\linewidth]{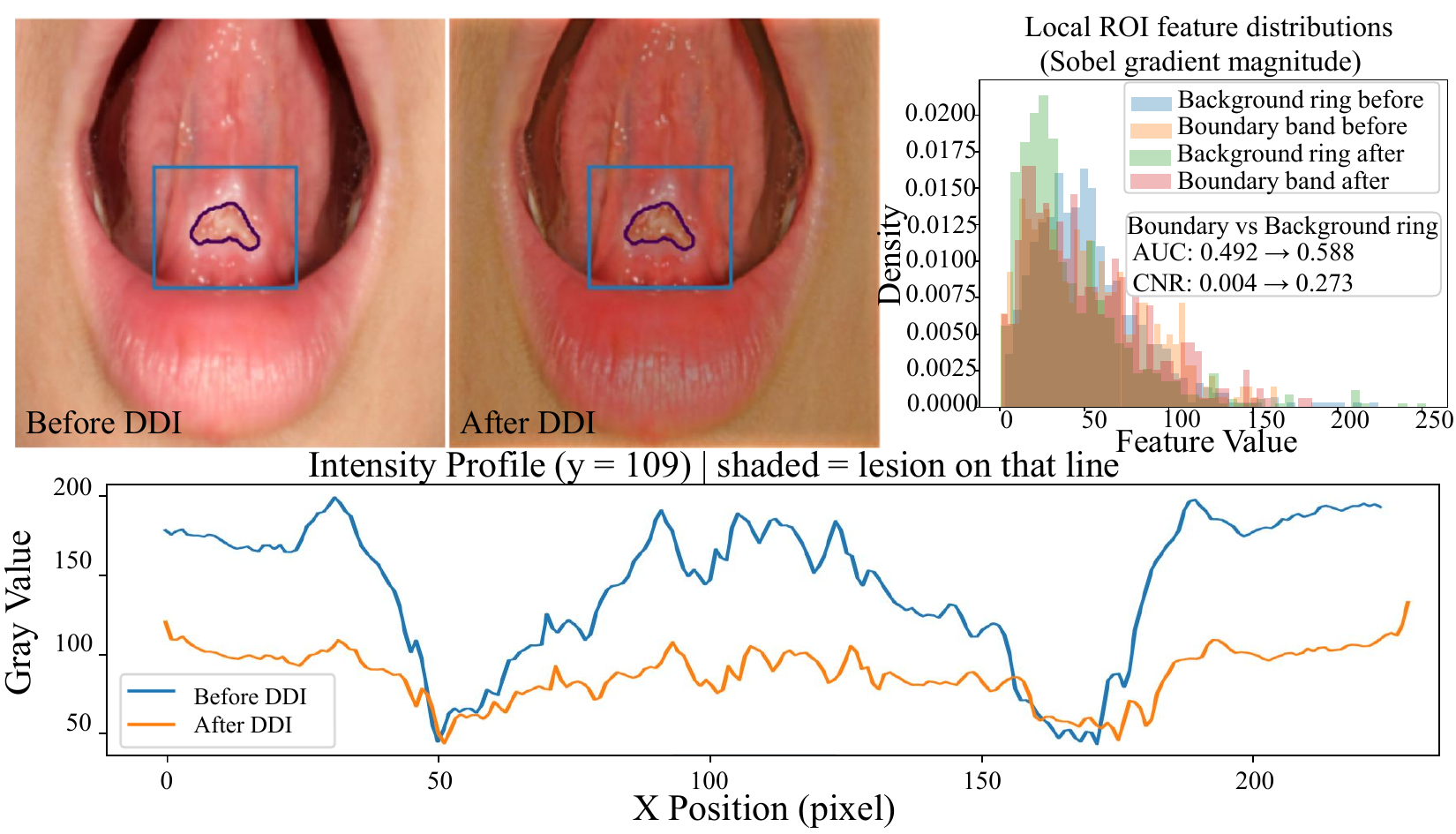}
\caption{Effectiveness of DDI. Comparison of local ROIs before and after DDI, along with Sobel gradient distributions and scanline intensity profiles. DDI suppresses spurious high-frequency responses while preserving true boundary structures, thereby improving lesion-background separability.}
\label{fig8}
\end{figure}

\subsubsection{DDI Module} 
As illustrated in \textcolor{blue}{Fig.}~\ref{fig8}, DDI modifies the local response distribution within the lesion-centered regions of interest (ROI). Compared with the response before DDI (top-left), the response after DDI exhibits reduced activation intensity in surrounding regions while preserving the main lesion contour. This effect is further reflected in the intensity profile along the sampled horizontal line (bottom), where the response becomes smoother and local fluctuations are noticeably attenuated.

The corresponding \textit{Sobel} gradient distributions (top-right) show improved separation between the lesion boundary band and the surrounding lesion-background ring after DDI, accompanied by consistent increases in \textit{Area-Under-the-Curve} (AUC) and Contrast-to-Noise Ratio (CNR). These results indicate that DDI suppresses degradation-sensitive and background-related responses while preserving boundary-relevant structures, leading to improved lesion-background separability and a more reliable feature representation for subsequent segmentation.
\begin{figure}[!t]
    \centering
    \includegraphics[width=0.98\linewidth]{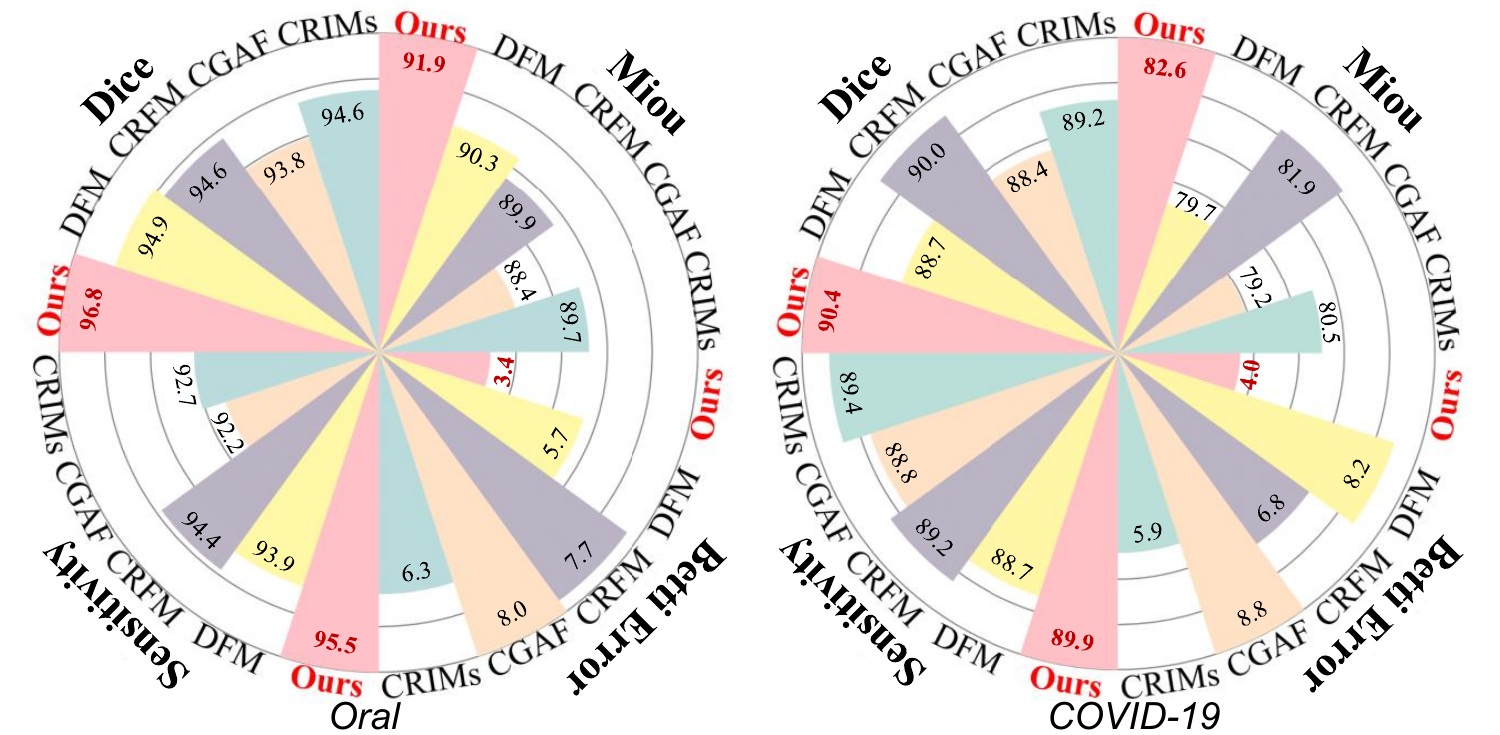}
    \caption{Effectiveness of CGF. Comparison with \textit{DFM}, \textit{CRIMs}, \textit{CRFM}, and \textit{CGAF} on \textit{Oral}   and \textit{COVID-19}. CGF achieves the best performance in both overlap accuracy and topological consistency.}
    \label{fig9}
\end{figure}

\subsubsection{CGF Module} 
As shown in \textcolor{blue}{Fig.}~\ref{fig9}, CGF is compared with several representative fusion modules, including \textit{DFM}~\cite{chen2026skin}, \textit{CRIMs}~\cite{chen2024feature}, \textit{CRFM}~\cite{yu2025stsnet}, and \textit{CGAF}~\cite{chen2024dea}. 
Specifically, \textit{DFM} performs pixel-wise adaptive fusion by concatenating two-branch features, \textit{CRIMs} uses channel-resolution cross-attention to exchange information between high-resolution location features and high-channel shape features,   \textit{CRFM} conducts cross-resolution feature interaction to align and fuse features with different spatial granularities, and \textit{CGAF} adopts content-guided attention to estimate spatial mixing weights and combines low- and high-level features. 
CGF consistently achieves the best overall performance. On the \textit{Oral} dataset, it achieves the highest \textit{Dice} (96.8), \textit{Sensitivity} (95.5), and \textit{mIoU} (91.9), while achieving the lowest \textit{Betti Error} (3.4). Similar trends are observed on the \textit{COVID-19} dataset, where CGF obtains the highest \textit{Dice} (90.4), \textit{Sensitivity} (89.9), and \textit{mIoU} (82.6), together with the lowest \textit{Betti Error} (4.0).

The consistent improvements across both overlap- and topology-related metrics indicate that CGF effectively enhances cross-backbone feature integration. By combining contour-guided selective interaction with local-consistency modulation, CGF better coordinates CNN-based local details and VSSM-based global structure, resulting in improved robust MIS performance.
\begin{figure}[!t]
    \centering
    \includegraphics[width=0.98\linewidth]{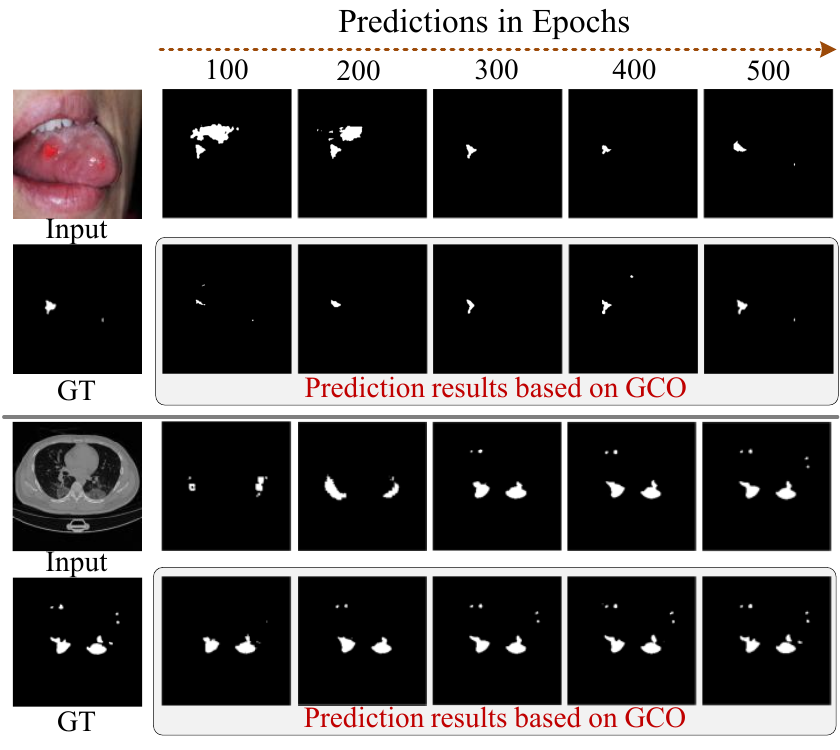}
\caption{Effectiveness of GCO. Prediction evolution across training epochs on representative samples from \textit{Oral} and \textit{COVID-19} datasets.} 
    \label{fig10}
\end{figure}

\subsubsection{GCO Module} 
As illustrated in \textcolor{blue}{Fig.}~\ref{fig10}, we visualize prediction evolution at 100, 200, 300, 400, and 500 epochs for representative \textit{Oral} and \textit{COVID-19} samples. Without GCO, the predictions exhibit noticeable instability across training epochs, characterized by fluctuations in lesion size and spatial localization. For the \textit{Oral} sample, lesion responses emerge at early epochs but progressively shrink from 300 to 500 epochs. Similarly, the \textit{COVID-19} sample shows irregular variations in both lesion extent and spatial distribution throughout training.
\begin{figure}[!t]
    \centering
    \includegraphics[width=8.8cm]{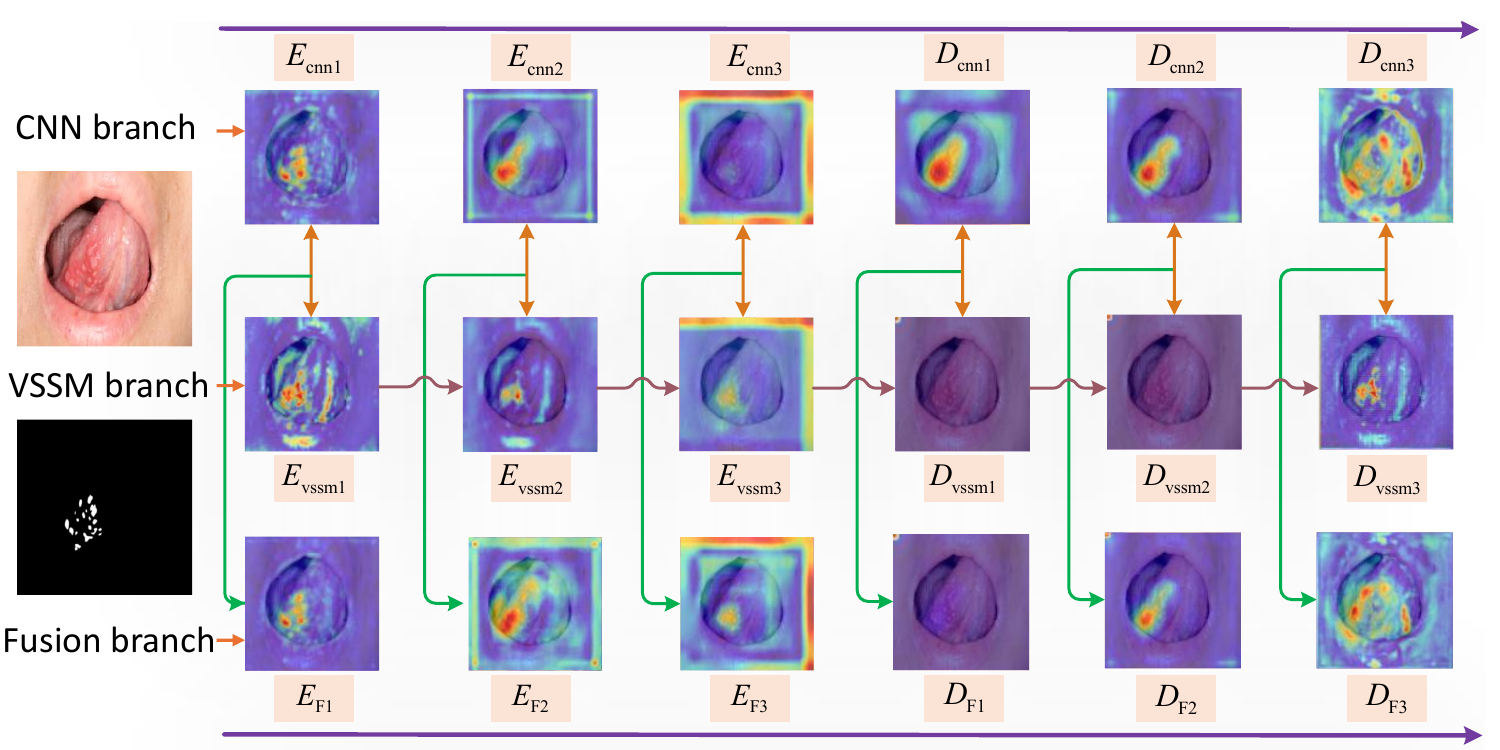}
    \caption{Activation responses of the proposed local–global decoupling and scale-wise interaction. Compared with the CNN and VSSM branches, the fused branch produces more complete and lesion-aligned activations during decoding, indicating effective integration of local detail and global structure.}
    \label{fig11}
\end{figure}

In contrast, incorporating GCO yields more stable prediction evolution. For the \textit{Oral} sample, the predicted lesion remains consistently localized around the GT region across all epochs, avoiding the abrupt shrinkage observed without GCO. For the \textit{COVID-19} sample, the predicted regions maintain more consistent bilateral localization and better preserve the major lesion structures throughout training. These observations suggest that GCO stabilizes the optimization trajectory through gradient-context-aware learning rate adaptation, resulting in more consistent convergence and reliable segmentation predictions.

\subsection{Activation Response Analysis}
As shown in \textcolor{blue}{Fig.~\ref{fig11}}, we visualize the activation responses of the CNN branch, the VSSM branch, and the fusion branch at the encoding and decoding stages. The three branches exhibit complementary response patterns. The CNN branch produces strong lesion-related activations in shallow encoding stages (\textit{e.g.}, $E_{\text{cnn1}}$ and $E_{\text{cnn2}}$) and decoding stages (\textit{e.g.}, $D_{\text{cnn1}}$ and $D_{\text{cnn2}}$), indicating high sensitivity to local textures, boundary cues, and fine structural variations. However, lesion-background and boundary-adjacent responses are also observed in deeper stages, suggesting susceptibility to local interference.
\begin{table}[!t]
\centering
\scriptsize
\setstretch{0.8}
\setlength{\tabcolsep}{6pt}
\renewcommand{\arraystretch}{1.08}
\caption{Computational complexity and inference efficiency under a unified input resolution and hardware setting. Lower values indicate better performance. Best results are highlighted in \textbf{bold}.}
\label{tab:complexity_comparison}
\begin{tabular}{lrrr}
\toprule
\textbf{Method} & \textbf{Params (M)$\downarrow$} & \textbf{GFLOPs $\downarrow$} & \textbf{Time (msec.)$\downarrow$} \\
\midrule

\textit{U$^2$-Net}         & 1.131   & 9.751    & 17.047 \\
\textit{DeepLabV3+}       & 40.347  & 13.268   & 7.433 \\
\textit{SFM}              & 25.874  & 7.871    & 19.055 \\
\textit{TherNet}          & 48.072  & 27.835   & 24.513 \\
\textit{MWCNN}            & 24.929  & 51.907   & 4.619 \\
\textit{MedSAM}           & 90.689  & 35.565   & 18.478 \\

\textit{BCDU-Net}         & 21.031  & 28.237   & \textbf{2.923} \\
\textit{MTG-Net}          & 26.992  & 151.940  & 12.563 \\
\textit{FAT-Net}          & 28.759  & 22.781   & 12.738 \\
\textit{MISSFormer}       & 42.462  & 9.860    & 22.679 \\
\textit{LeViT-UNet}       & 52.154  & 25.547   & 13.141 \\
\textit{Swin-UNet}        & 27.168  & 6.126    & 10.134 \\
\textit{CyCoSeg}          & 0.053   & 0.060    & 17.165 \\
\textit{Mamba-UNet}       & 44.274  & 7.517    & 29.675 \\
\textit{UltraLight VM}    & \textbf{0.049} & \textbf{0.052} & 14.942 \\
\textit{LKM-UNet}         & 15.317  & 21.778   & 22.174 \\
\textit{H-vmuNet}         & 8.973   & 0.808    & 79.417 \\
\textit{Mamba-Sea}        & 27.447  & 4.091    & 19.659 \\
\textit{FusionNet}        & 4.918   & 5.149    & 6.306 \\
\textit{MBP-SSNet}        & 639.272 & 80.752   & 11.899 \\
\textit{EAG-SAE-ESTF}     & 48.843  & 36.381   & 10.305 \\
\textit{GobletNet}        & 133.248 & 141.612  & 8.366 \\
\textit{MADGNet}          & 31.219  & 7.666    & 32.389 \\
\textit{FFTMed}           & 3.919   & 1.659    & 12.379 \\
\textit{CSWin-UNet}       & 23.568  & 5.009    & 33.237 \\
\textit{SAM2-UNet}        & 216.530 & 93.961   & 28.169 \\

\textit{MedT}             & 11.354  & 18.268   & 13.535 \\
\textit{TransFuse}        & 14.637  & 5.904    & 8.172 \\
\textit{H2Former}         & 33.679  & 24.824   & 13.734 \\
\textit{MambaEviScrib}    & 3.706   & 4.215    & 13.043 \\
\textit{XNet}             & 135.816 & 114.736  & 6.685 \\
\textit{DS2F}             & 39.205  & 36.991   & 32.385 \\

\rowcolor{black!6}
Ours             & 158.310  & 12.533   & 37.902 \\
\bottomrule
\end{tabular}
\end{table}

In contrast, the VSSM branch generates more spatially distributed activations, reflecting its ability to capture broader contextual information. Nevertheless, its responses around lesion regions are occasionally less concentrated, as observed in $D_{\text{vssm1}}$ and $D_{\text{vssm2}}$, indicating reduced sensitivity to small or localized structures.

The fusion branch (see \textcolor{blue}{Fig.~\ref{fig2}}) effectively combines these complementary characteristics. Across both the encoding and decoding stages, lesion-related activations are preserved more consistently (\textit{e.g.}, $E_{\text{F1}}$–$E_{\text{F3}}$ and $D_{\text{F2}}$–$D_{\text{F3}}$), while irrelevant lesion-background responses are progressively suppressed. As a result, the fused features exhibit stronger alignment with the lesion regions, demonstrating that the proposed fusion strategy successfully integrates CNN-based local discriminability with VSSM-based contextual modeling to provide more informative representations for segmentation.

\subsection{Computational Complexity}
\textcolor{blue}{Table}~\ref{tab:complexity_comparison} compares model complexity and inference efficiency in terms of parameter count, GFLOPs, and per-image latency. A clear trade-off can be observed among existing methods. Lightweight models such as \textit{CyCoSeg} and \textit{UltraLight VM} achieve low computational cost, but their limited capacity may constrain performance under heterogeneous imaging conditions. Conversely, high-capacity models, including \textit{XNet}, \textit{GobletNet}, and \textit{MBP-SSNet}, require substantially more parameters or computational operations, indicating that increasing model scale alone does not necessarily yield a more efficient or effective solution.

\ourMethod{} contains 158.31M parameters, 12.533 GFLOPs, and 37.902 millisecond latency per image. Although the dual-backbone architecture and cross-backbone interactions increase the parameter count and inference latency, the overall computational cost remains competitive, with lower GFLOPs than several high-capacity baselines, including \textit{MTG-Net}, \textit{MBP-SSNet}, \textit{GobletNet}, and \textit{XNet}. Combined with the quantitative segmentation results, these findings suggest that \ourMethod{} achieves a favorable balance between segmentation performance and computational efficiency, with its gains arising from effective architectural design rather than model scale alone.

\subsection{Further Cross-Architecture Validation of GCO}
\begin{figure}[!t]
    \centering
    \includegraphics[width=0.98\linewidth]{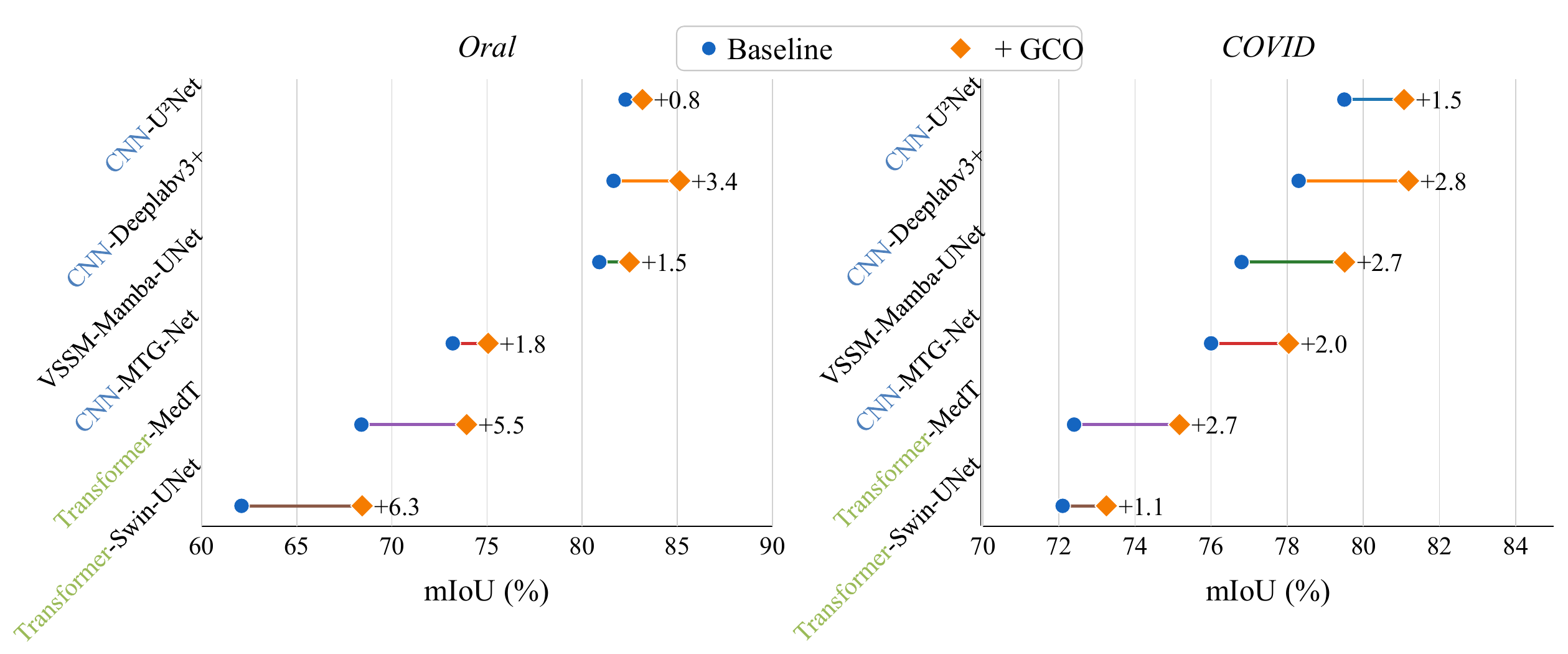}
\caption{Cross-backbone generalization of GCO on the \textit{Oral} and \textit{COVID-19} datasets.}
    \label{fig12}
\end{figure}

To evaluate the generalization of GCO beyond \ourMethod{}, we integrate it into six representative segmentation backbones spanning CNN-, Transformer-, and VSSM-based architectures. As shown in \textcolor{blue}{Fig.}~\ref{fig12}, GCO consistently improves \textit{mIoU} on both the \textit{Oral} and \textit{COVID-19} datasets across all evaluated models, yielding average gains of 3.2 and 2.1 points, respectively. These consistent improvements suggest that the effectiveness of GCO is not tied to a specific network architecture.

Combined with the prediction evolution analysis in \textcolor{blue}{Fig.}~\ref{fig10}, the results indicate that GCO primarily operates at the optimization level rather than the representation level. By adaptively regulating the learning rate based on gradient-context information, GCO promotes more stable training dynamics and reliable convergence, leading to improved segmentation performance across diverse backbone architectures and heterogeneous medical imaging scenarios.

\section{Conclusion and Future Work}
This paper presents an optimization-aware dual-domain network for robust medical image segmentation (\ourMethod{}) under composite degradations and multi-scale geometric variations. The proposed framework is based on the observation that degradation suppression, local-global representation learning, and optimization stability are inherently coupled and should be addressed in a coordinated manner. To this end, \ourMethod{} integrates dual-domain feature refinement, contour-guided cross-backbone interaction, and gradient-context-aware optimization within a unified framework. Extensive experiments on six benchmark datasets spanning multiple imaging modalities and anatomical targets demonstrate consistent improvements in segmentation accuracy, sensitivity, and topological consistency, highlighting the effectiveness of jointly modeling feature representation and optimization dynamics.

Despite these promising results, several limitations remain. The current study is restricted to 2D segmentation tasks, and extending the framework to 3D volumes and temporal imaging data requires further investigation. In addition, resizing inputs to a fixed resolution may compromise extremely fine-grained structures, while the dual-backbone design introduces additional parameters and inference latency compared with lightweight architectures. Although improved topological metrics indicate enhanced structural consistency, further clinical validation is necessary to assess practical utility in real-world settings. Future work will focus on extending the framework to 3D and longitudinal imaging scenarios, improving robustness under limited annotations and cross-center domain shifts, and incorporating uncertainty-aware modeling to further enhance reliability and generalization.


\ifCLASSOPTIONcaptionsoff
  \newpage
\fi

\bibliographystyle{IEEEtran}
\bibliography{medNet-2026}


\end{document}